\documentclass[10pt,twocolumn,letterpaper]{article}
\usepackage[pagenumbers]{cvpr}
\usepackage{newtxmath}
\usepackage{fontspec}
\usepackage{bbm}
\graphicspath{{figures/}}

\usepackage{placeins}
\usepackage{tcolorbox}

\usepackage[breaklinks,hidelinks]{hyperref}

\title{Towards Embodied Air-Ground Cooperative Object Search:\\Benchmark, Dataset and Agentic Method}
\author{
Boao Yu$^{1,2}$, Zimo Chen$^{1,2}$, Junreng Rao$^{1,2}$, Yue Hu$^{1,2}$, Zhengqiu Zhu$^{1,2}$,\\
Yong Zhao$^{1,2}$, Jianzhe Gao$^{3}$, Pengfei Zhu$^{4}$, Rusheng Ju$^{1,2}$\\[0.45em]
{\small $^{1}$College of Systems Engineering, National University of Defense Technology}\\
{\small $^{2}$National Key Laboratory of Digital Intelligent Modeling and Simulation}\\
{\small $^{3}$Zhejiang University\qquad $^{4}$Tianjin University}
}

\begin{document}

\twocolumn[{
\begin{@twocolumnfalse}
\maketitle
\begin{center}
\includegraphics[width=0.86\textwidth]{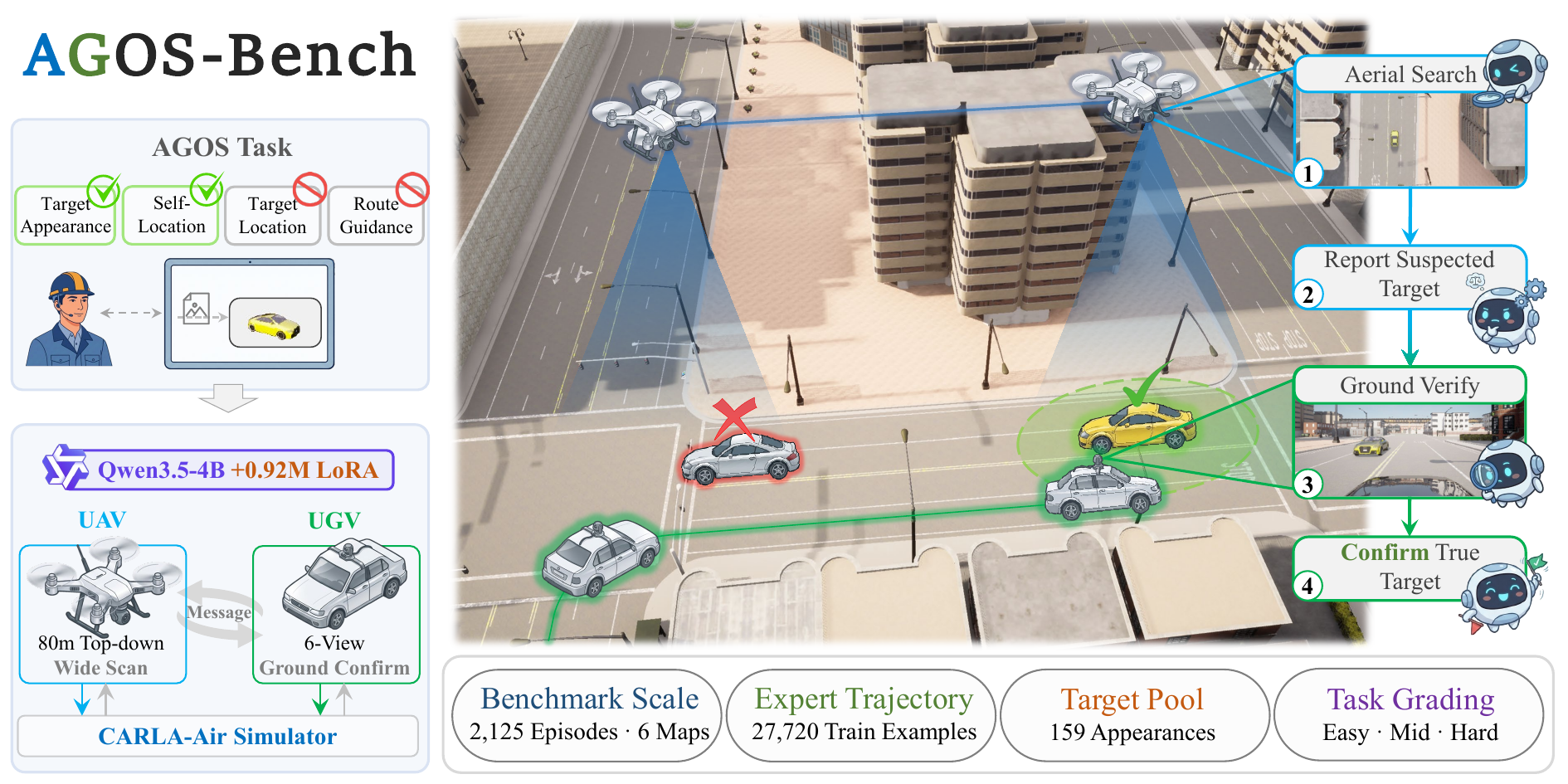}
\captionof{figure}{Overview of the AGOS-Bench. An episode succeeds only when the UGV stops within 30\,m of the target with line of sight, shows the target in its six views, and issues a MATCH verdict.}
\label{fig:teaser}
\end{center}
\end{@twocolumnfalse}
}]

\begin{abstract}
Urban object search, locating a specific object from a single reference image without route or direction instructions and confirming the instance up close, is a fundamental capability for embodied agents. Existing urban embodied benchmarks evaluate one segment of this chain: navigation ends at a coordinate, and detection and question answering stop at an answer. We present AGOS-Bench, the first benchmark for urban cooperative object search by an air-ground team, integrating aerial discovery and ground-level instance verification in one closed loop. Nine metrics assess search, cooperation, and verification, and they distinguish ``found but unconfirmed'' from ``never found''. AGOS-Dataset instantiates the task in 2{,}125 episodes across six city maps under seen and unseen test settings, contributing 27{,}720 Train Examples along with 1{,}100 Expert Trajectories. To solve the task, we propose AGOS-Agent, which combines algorithmic coverage and approach routing with event-driven MLLM judgments for target verification and cooperation. On the unseen test set, zero-shot AGOS-Agent (Qwen3.5-4B) achieves a success rate (SR) of 37.67\%, 7.00 percentage points (pp) above a controlled baseline with identical information, actions, and prompts. LoRA fine-tuning raises SR to 57.00\% (+19.33 pp). Adapted existing systems AGC-VLN-OS and AeroDuo-OS achieve SRs of 31.33\% and 28.67\%, respectively. Project page: \url{https://github.com/Yomier/AGOS-Bench}.
\end{abstract}

\section{Introduction}
\label{sec:intro}

Urban object search is an important problem in embodied intelligence, with direct applications in emergency response, security patrol, and post-disaster rescue. In these scenarios, a specific object must be located based on a single visual reference within a city district and approached for close-range instance confirmation. A ground robot can verify a target at close range, but street-level views limit its search coverage. A single aerial robot sweeps a district quickly, yet its top-down view lacks the close-range detail needed to tell similar objects apart. An unmanned aerial vehicle (UAV) and an unmanned ground vehicle (UGV) therefore form a complementary team, where the UAV discovers over a wide area and the UGV verifies up close.

Existing urban benchmarks study perception, search, and navigation under different information conditions (Table~\ref{tab:scope}). AirCopBench~\cite{zha2025aircopbench} evaluates perception and question answering from synchronized multi-UAV views, while CityAVOS~\cite{ji2026cityavos} studies object search with a single UAV. AirGroundBench~\cite{li2026airgroundbench} combines aerial-ground question answering with navigation episodes, and AirGroundVLN~\cite{zeng2026airgroundvln} evaluates cooperative navigation toward language-specified targets with coarse directional descriptions. These settings do not jointly evaluate an air-ground team searching without target-location or route hints and confirming the target from ground observations. Such searches often begin with an incidentally captured image as the only target-specific information.

To bridge this gap, we formulate the embodied Air-Ground Object Search (AGOS) task and introduce AGOS-Bench. A UAV-UGV team receives a region map and public road topology and cooperatively locates a target specified by a single reference image. Completion requires the UGV to approach and confirm the target from ground observations, linking a visual match to the location needed for subsequent on-site inspection or action. Agents must therefore coordinate exploration under a limited budget and resolve ambiguous matches across aerial and ground views. To support evaluation and learning under this setting, AGOS-Dataset supplies parked-vehicle search episodes and Expert Trajectories. The resulting Train Examples pair observations with action or verification labels using the same interfaces as online evaluation. AGOS-Agent provides a reference solution through an algorithm-and-tool layer and a learnable multimodal large language model (MLLM) layer. Algorithms coordinate coverage and approach routing, while frozen pretrained perception tools detect candidates and score their similarity to the reference. Candidate evidence and task events trigger MLLM judgments about target matches and handoff requests, leaving routine movement to algorithms. Both members share the same model with role-specific prompts.

Our main contributions are:
\begin{itemize}
\setlength{\itemsep}{0pt}\setlength{\parskip}{0pt}\setlength{\parsep}{0pt}
  \item \textbf{Task and benchmark.} We formulate the AGOS task and introduce AGOS-Bench, the first benchmark for air-ground cooperative object search without target-location or route hints. Success requires ground-level target confirmation, and nine metrics assess search, cooperation, verification, and decision efficiency.
  \item \textbf{Dataset and training resource.} We release AGOS-Dataset with 2{,}125 episodes across six city maps, divided into training, validation, seen-test, and unseen-test splits. The 1{,}100 training episodes provide one Expert Trajectory each, yielding 27{,}720 Train Examples.
  \item \textbf{Method and findings.} AGOS-Agent pairs deterministic planning with sparse MLLM wakeup. On the unseen-city test set, it improves SR over the controlled baseline on seven of nine backbones and reduces MLLM calls on all nine. With Qwen3.5-4B, LoRA fine-tuning with Train Examples further raises SR from 37.67\% to 57.00\%. Comparatively, adapted existing systems AGC-VLN-OS~\cite{zhang2026agcvln} and AeroDuo-OS~\cite{wu2025aeroduo} attain SRs of 31.33\% and 28.67\%, respectively.
\end{itemize}

\begin{table}[t]
  \centering
  \footnotesize
  \setlength{\tabcolsep}{1pt}
  \caption{Capability comparison with representative benchmarks. ``No route'': no step-by-step path instruction is given. ``No target pos.'': neither the target's position nor a direction hint is given. ``Ground confirm.'': the success criterion itself requires a stationary ground instance verdict, beyond any method-internal check. ``--'' marks inapplicable dimensions.}
  \label{tab:scope}
  \resizebox{\columnwidth}{!}{%
  \begin{tabular}{@{}lllccc@{}}
    \toprule
    Work & Task type & Config. & No route & No target pos. & Ground confirm. \\
    \midrule
    CityAVOS~\cite{ji2026cityavos} & Object search & UAV-only & $\checkmark$ & $\checkmark$ & $\times$ \\
    UAV-ON~\cite{xiao2025uavon} & Object navigation & UAV-only & $\checkmark$ & $\checkmark$ & $\times$ \\
    CityEQA~\cite{zhao2025cityeqa} & Embodied QA & UAV-only & $\checkmark$ & $\times$ & $\times$ \\
    CityCube~\cite{xu2026citycube} & Cross-view reasoning & -- & -- & -- & $\times$ \\
    AirCopBench~\cite{zha2025aircopbench} & Multi-UAV perception & Multi-UAV & -- & -- & $\times$ \\
    AirGroundBench~\cite{li2026airgroundbench} & QA and navigation & UAV+UGV & $\times$ & $\times$ & $\times$ \\
    AirGroundVLN~\cite{zeng2026airgroundvln} & Air-ground VLN & UAV+UGV & $\checkmark$ & $\times$ & $\times$ \\
    AGC-VLN~\cite{zhang2026agcvln} & Air-ground VLN & UAV+UGV & $\checkmark$ & $\times$ & $\times$ \\
    \midrule
    \textbf{AGOS-Bench} & \textbf{Air-ground object search} & \textbf{UAV+UGV} & \textbf{$\checkmark$} & \textbf{$\checkmark$} & \textbf{$\checkmark$} \\
    \bottomrule
  \end{tabular}}
\end{table}

\section{Related Work}
\label{sec:related}

We review three lines of work: benchmarks and simulation platforms for AGOS-Bench, search and cooperation methods for AGOS-Agent, and demonstration datasets for AGOS-Dataset.

\begin{figure*}[!t]
  \centering
  \includegraphics[width=0.95\textwidth]{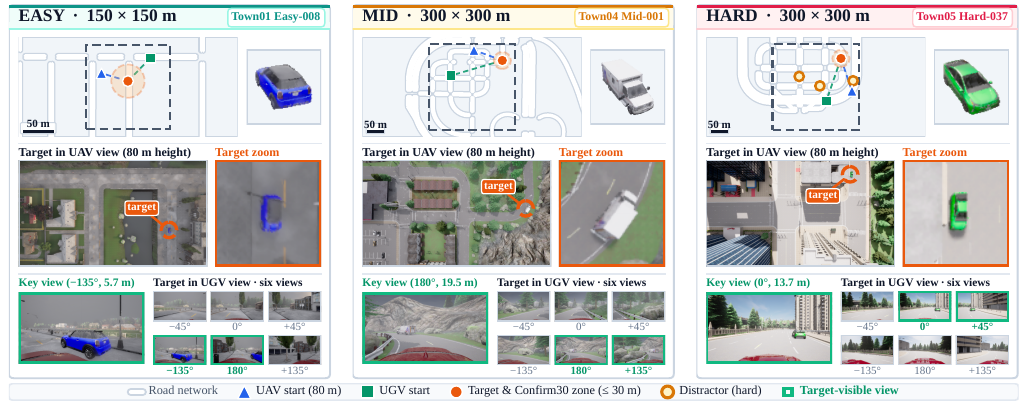}
  \caption{Task examples at three difficulty levels (one episode per level). Top row: the search region with the start points of the team, the target (orange dot with its confirmation zone, $\le30$\,m), look-alike distractors (orange hollow dots, hard only), and the target appearance reference. Middle row: the target in the UAV top-down view at 80\,m, with a zoomed crop. Bottom row: the key view (the six-view direction facing the target, labeled with its azimuth and the distance to the target) and the six-view set from the parked UGV at the ground confirmation moment. Green boxes mark the views in which the target is visible.}
  \label{fig:task-examples}
\end{figure*}

\textbf{Benchmarks and simulation platforms.} Embodied benchmarks such as Habitat~\cite{savva2019habitat} and AI2-THOR~\cite{kolve2017ai2thor} standardize observations, actions, and success metrics, mostly indoors. Simulators have reached city scale~\cite{dosovitskiy2017carla,gao2024embodiedcity,wu2025metaurban,zhuang2025simworld,liu2023aerialvln,lee2024citynav,gao2025openfly,yao2026aeroverse}, and air-ground unified simulators~\cite{cui2026airsimag,garimella2026hercules,wang2025transimhub,zeng2026carlaaAir} share one physics clock across both members. Among city-scale benchmarks, AirCopBench~\cite{zha2025aircopbench} scores 14{,}610 questions over 2{,}920 synchronized multi-UAV view sets without embodied control. AirGroundBench~\cite{li2026airgroundbench} adds 115 closed-loop navigation episodes to 62k cross-view questions over 1{,}021 synchronized aerial-ground view sets. AirGroundVLN~\cite{zeng2026airgroundvln} sends an air-ground team to a language-specified target with a coarse relative direction and environment context, scoring arrival within 5~m. City-scale QA, cross-view reasoning, single-aerial search, and video understanding cover the remaining subproblems~\cite{zhao2025cityeqa,zhou2025urbench,xu2026citycube,ji2026cityavos,xiao2025uavon,wang2025cityseeker,zhao2025urbanvideo}. AGOS-Bench combines these capabilities in an air-ground search task with unknown target location and explicit ground-level instance confirmation.

\textbf{Search and cooperation methods.} Search methods rank candidate locations using visual similarity and language priors~\cite{gadre2023cows,yu2023l3mvn}, or predict likely goals with world models and fine-tuned vision-language models~\cite{nie2025wmnav,yokoyama2025filmnav}. Panoramic parsing provides visual context~\cite{jin2025panonav}, while end-to-end models use observation histories to select actions~\cite{zhang2024navid,zhang2024uninavid,wang2025skyvln,xu2025geonav}. Cooperative methods address task allocation and communication across multiple robots~\cite{wu2025aeroduo,song2026conavuav,wang2026colasn,huang2025kiterunner,zhang2026agcvln,wang2025rally,yu2024conavgpt,rajvanshi2025sayconav}. The closest air-ground navigation systems still supply target-side priors, including coarse directions in AirGroundVLN~\cite{zeng2026airgroundvln} and goal positions in AGC-VLN~\cite{zhang2026agcvln}. Their success criteria measure stopping distance to the target. Cross-view re-identification studies appearance matching across aerial and ground views~\cite{shang2025agcvreid,li2025gsalign,yin2024crossview}. AGOS-Agent combines algorithmic coverage and routing with MLLM-based appearance judgments throughout cooperative search and confirmation.

{\setlength{\parskip}{0pt}
\textbf{Demonstration data and policy tuning.} R2BC~\cite{mattson2025r2bc} learns multi-agent behavior from single-agent demonstrations, RoboTalk~\cite{goldfajn2026robotalk} learns multi-robot communication and coordination from multimodal demonstrations, MATE~\cite{yu2026mate} collects collaboration data on a virtual teleoperation platform, and CHORUS~\cite{doshi2026chorus} shows that one policy conditioned on embodied identities can serve heterogeneous robots. Reinforcement tuning improves cooperative vision-language-action models~\cite{xu2027coopvla}; Observe-Share-Predict~\cite{szlatincsan2026observe} dissects shared observations and teammate prediction. The demonstrations behind these works are either collected through human teleoperation or recorded over low-level actions end to end. AGOS-Dataset supplies reproducible heterogeneous cooperation demonstrations whose training interfaces match online inference.\par}

\section{AGOS-Bench}
\label{sec:bench}

\subsection{Task Formulation}

AGOS-Bench formalizes air-ground cooperative object search in urban environments: within a given town region, a team of one UAV and one UGV must locate a target vehicle $o$ from a single reference image $I_{\text{ref}}$ and confirm on the ground that it matches the reference. At the start of each episode the team receives the region map $M$, the public road topology, and both initial poses $P^{r}_{0}$. The true position of $o$ remains unknown. Each member selects and executes actions on its own schedule. A task step executes the next action of whichever member becomes available first, with availability determined by the durations of its preceding actions. At step $t$, the selected role $r\in\{\text{UAV},\text{UGV}\}$ chooses an action based on its search history $\mathcal{H}^{r}_{t}$ and the text-only cooperation message $T_{t}$, with superscript $r$ marking the role throughout:
\begin{equation}
  a^{r}_{t} = \pi^{r}\!\left(M,\; I_{\text{ref}},\; \mathcal{H}^{r}_{t},\; T_{t}\right),\qquad
  a^{r}_{t}\in\mathcal{A}^{r}.
  \label{eq:task}
\end{equation}
Here $\mathcal{A}^{r}$ denotes the discrete action set available to role $r$. The two members act asynchronously, allowing search and verification to overlap and the UGV to find the target independently.

\subsection{Inputs, Observations, and Actions}

The allowed priors are public GPS, road topology, camera geometry, and the agent's own pose, observation history, executed actions, and cooperation messages. No oracle information of any form is provided: no target position or orientation, no distribution of vehicles in unobserved areas, no city-wide top-down image given in advance, and no distance or direction hints. The UAV observes a top-down RGB image and the UGV a six-view RGB set, with the resolutions and camera geometry given in Appendix~\ref{sec:appendix-actions}. Both members act through the discrete atomic actions of Eq.~\eqref{eq:task}: the UAV moves between grid waypoints in eight horizontal directions, annotates candidates, requests assistance, or pauses to observe, and the UGV moves forward along roads, reverses, turns at intersections, confirms, rejects, or stops to observe. Movement cost is measured by path length, and virtual time advances with action duration, so the two members decide asynchronously.

\subsection{Success Criterion and Scoring}

An episode succeeds only when the following four conditions hold jointly:
\begin{equation}
\begin{aligned}
  \text{success} \;=\;
  &\mathbbm{1}\!\left[d(\text{UGV},o)\le 30\,\text{m}\right]\cdot
  \mathbbm{1}\!\left[\text{LOS}\right]\\
  &{}\cdot\mathbbm{1}\!\left[\mathrm{Vis}(o,\,V_{\text{stop}})\right]\cdot
  \mathbbm{1}\!\left[v=\text{MATCH}\right],
\end{aligned}
\label{eq:confirm30}
\end{equation}
where $d$ denotes the horizontal distance between the UGV and the target. LOS requires an unobstructed line of sight at the stopping pose, whereas $\mathrm{Vis}$ additionally requires the target to appear in the six-view observations. The verdict $v$ is MATCH, NON\_MATCH, or UNKNOWN, indicating a target match, a mismatch, or insufficient evidence. At ground confirmation, only a MATCH verdict can confirm the target. NON\_MATCH and UNKNOWN release the candidate and resume the search. An episode fails after a false confirmation, exhaustion of the 80-step budget, or a method-side stall. Site-specific verdict behavior is detailed in Appendix~\ref{sec:appendix-verdicts}.

\begin{figure*}[t]
  \centering
  \includegraphics[width=\textwidth]{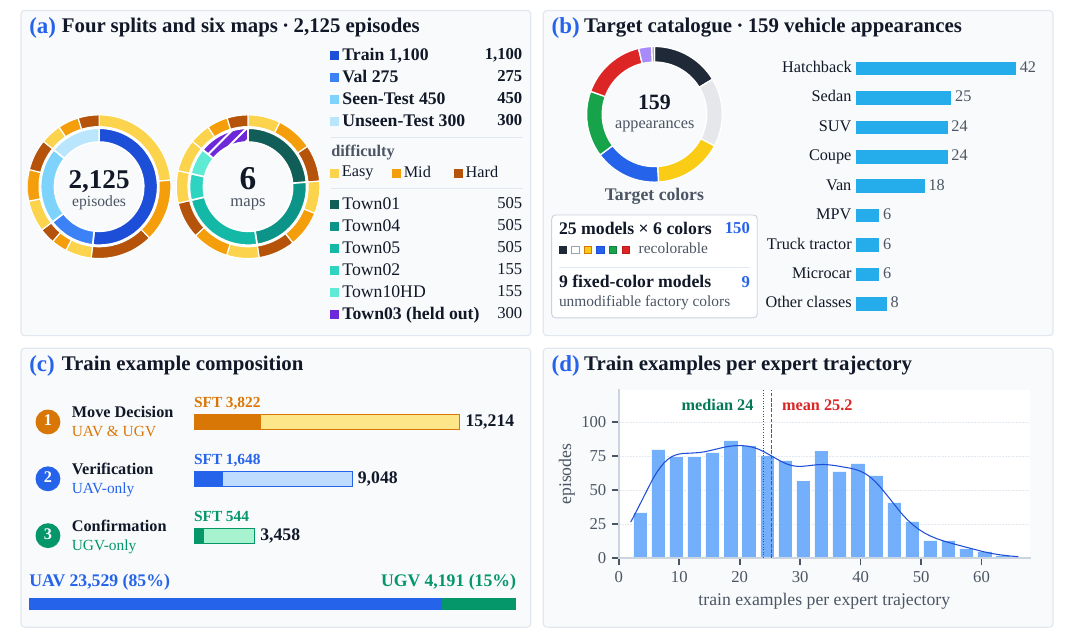}
  \caption{Overview of AGOS-Dataset. (a) The 2{,}125 episodes by subset and by map, with Town03 as the only held-out map. (b) The 159 vehicle appearances by color and vehicle class. (c) Released and SFT-used Train Examples by category, with UAV/UGV counts. (d) Train Examples per Expert Trajectory (mean 25.2, median 24).}
  \label{fig:dataset}
\end{figure*}

\subsection{Evaluation Metrics}
\label{sec:metrics}

We report nine metrics, including four adapted from conventional navigation evaluation. \textbf{SR} (Success Rate) is the fraction of episodes ending in a correct confirmation, and \textbf{OSR} (Oracle Success Rate) counts episodes in which the UGV reaches a pose that satisfies the oracle conditions for initiating confirmation. The OSR--SR gap measures the remaining loss between legal ground reach and successful confirmation. \textbf{SPL} (Success weighted by Path Length) weighs each success by path efficiency, and \textbf{NE} (Navigation Error) measures the final distance to the target.

\textbf{ASE} (Aerial Search Efficiency) measures how early the UAV obtains a legal view of the target, using UAV-side decision steps. \textbf{ACG} (Aerial Candidate Guidance) gives higher scores to candidates that are closer to the target and nominated earlier. ASE evaluates aerial discovery, while ACG evaluates the location guidance passed to the ground member. \textbf{GVA} (Ground Verification Accuracy) measures the correctness of ground confirmation and rejection decisions at the event level.

\textbf{DS} (Decision Steps) is the average number of MLLM calls per episode. \textbf{DSR} (Decision-Step-Weighted Success Rate) weights each successful episode by the reciprocal of its call count. In the main-text tables, normalized metrics are scaled by 100, NE is reported in meters, and DS is measured in calls per episode. Detailed definitions and calculations are given in Appendix~\ref{sec:appendix-metrics}.

\section{AGOS-Dataset}
\label{sec:dataset}

AGOS-Dataset is collected in CARLA-Air~v0.1.7~\cite{zeng2026carlaaAir} across six city maps, Town01, Town02, Town03, Town04, Town05, and Town10HD. It contains 2{,}125 episodes in four disjoint splits: 1{,}100 training, 275 validation, 450 seen-test, and 300 unseen-test episodes (Figure~\ref{fig:dataset}(a)).

\begin{figure*}[t]
  \centering
  \includegraphics[height=241.982606bp]{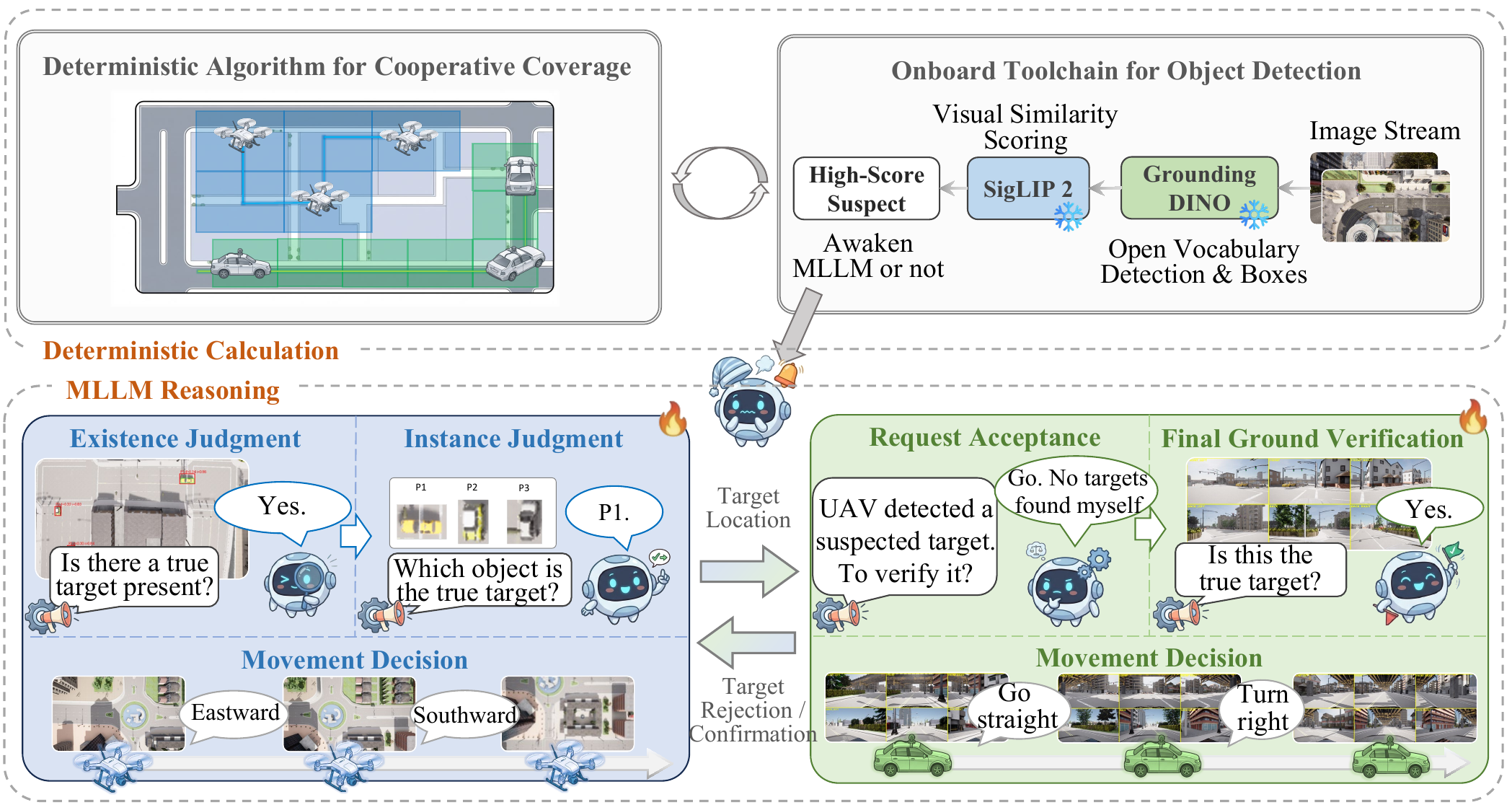}
  \caption{The two-layer AGOS-Agent framework. The algorithm-and-tool layer (top) combines cooperative coverage with frozen detection and similarity models that trigger MLLM calls. The learnable MLLM layer (bottom) serves the UAV (left) and UGV (right), coupled by target-location handoff and rejection or confirmation messages.}
  \label{fig:framework}
\end{figure*}

The training, validation, and seen-test splits share five cities, while Unseen-Test300 uses the held-out Town03 environment. All splits draw target vehicles from the same catalog of 34 models and 159 appearances (Figure~\ref{fig:dataset}(b)). Map-wise counts are given in Appendix~\ref{sec:appendix-maps}.

An algorithmic expert collects one team-level Expert Trajectory per training episode using public map geometry for search and approach planning. Ground truth supplies labels for candidate nomination, acceptance, and confirmation. Motion planning and model inputs use public information. These labels determine whether a candidate is handed over, approached, or confirmed during collection. Each Train Example pairs an instruction and observation with a structured action or verdict label at an expert decision point. The release also includes the camera frames and executed actions of both members within each Expert Trajectory. Labeling rules and quality checks are given in Appendices~\ref{sec:appendix-corpus} and~\ref{sec:appendix-corpus-quality}.

The training split provides 27{,}720 Train Examples from 1{,}100 Expert Trajectories in three categories. \textbf{Move Decision} contains routine search and recovery decisions from both members. \textbf{Verification} groups UAV first-check decisions with the crop-selection calls recorded at the same task step. \textbf{Confirmation} includes UGV candidate acceptance, approach-related decisions, inspection, and final confirmation. Their counts are 15{,}214, 9{,}048, and 3{,}458, respectively (Figure~\ref{fig:dataset}(c)). Figure~\ref{fig:dataset}(d) shows Train Examples per Expert Trajectory. The fine-tuning input comprises 8{,}212 Train Examples collected from 301 training episodes. Sequence-length filtering retains 6{,}014 of them for training. Detailed classification rules and counts are given in Appendices~\ref{sec:appendix-corpus} and~\ref{sec:appendix-corpus-quality}.

\textbf{Task difficulty.} We define three levels through search-region geometry and the strength of visual distractors, illustrated in Figure~\ref{fig:task-examples}. Easy settings use a $150\times150$\,m region with no distractor vehicles and place the target at least 10\,m from either start point. Mid settings expand the region to $300\times300$\,m and increase the minimum start distance to 100\,m. Hard settings retain the mid geometry and add two to three look-alike distractors, mostly the same vehicle model in different colors. Appendix~\ref{sec:appendix-maps} gives the three-level distractor selection rule. Easy episodes cover all six maps, while mid and hard episodes use Town01, Town03, Town04, and Town05. Easy tests completion of the basic search loop, mid tests large-area search, and hard tests search and verification under distraction.

\begingroup
\setlength{\parskip}{0pt}
\patchcmd{\cvprsection}{-10pt plus -2pt minus -2pt}{-10pt}{}{\errmessage{Section spacing patch failed}}
\section{AGOS-Agent}
\label{sec:agent}

AGOS requires broad-area exploration, cross-view instance recognition, and coordinated ground inspection. Deterministic algorithms provide stable, efficient coverage and routing from map geometry. Specialized perception tools turn image streams into candidate objects and reference-similarity scores. Ambiguous views and look-alike objects call for flexible instance judgments, which MLLMs can learn from demonstrations. Their inference cost and variable reliability make selective invocation central to the design. AGOS-Agent therefore combines two layers: algorithms and tools sustain the search, while an event-driven MLLM handles visual judgments and cooperation decisions (Figure~\ref{fig:framework}).\par
\endgroup

\subsection{Search with Algorithms and Perception Tools}

Joint coverage divides road-network coverage between UAV observation stations and UGV routes. Let $\mathcal{V}_{\text{A}}$ and $\mathcal{V}_{\text{G}}$ denote the units assigned to the two members. The planner aims to minimize the larger of the two estimated completion times over feasible assignments of all remaining coverage units:
{\small
\setlength{\jot}{1pt}
\begin{equation}
\begin{aligned}
&(\mathcal{V}_{\text{A}}^\star, \mathcal{V}_{\text{G}}^\star)
  = \operatorname*{arg\,min}_{(\mathcal{V}_{\text{A}}, \mathcal{V}_{\text{G}})}\\
&\quad \max \Big\{\hat{T}_{\text{UAV}}\big(\mathcal{V}_{\text{A}} \cup \mathcal{V}_{\text{aerial-only}},\, P_t^{\text{UAV}}\big),\\
&\phantom{\quad \max \Big\{}\hat{T}_{\text{UGV}}\big(\mathcal{V}_{\text{G}},\, P_t^{\text{UGV}}\big) \Big\},
\end{aligned}
\end{equation}
}
where $\hat{T}$ estimates completion time from the current pose $P_t^r$, and $\mathcal{V}_{\text{aerial-only}}$ contains units coverable only from the air. The algorithm first assigns these units to the UAV, then greedily allocates the rest using incremental flight distance and directed road distance. The UAV follows a sweep through observation stations, while the UGV follows shortest legal road routes. Replanning uses current positions and preserves completed coverage when a request is accepted or verification ends.

Both members use Grounding DINO to obtain candidate vehicle crops and SigLIP 2 to compute cosine-based similarity scores between each crop and the reference image. For the UAV, new evidence with an above-threshold similarity score triggers an MLLM judgment. Requests and candidate arrivals also trigger calls. Duplicate suppression avoids repeated calls on unchanged evidence. Coverage and approach routing continue algorithmically between calls. Baseline-Agent instead uses fixed coverage and calls the MLLM at coverage decision points, retaining the same perception and verification components. Details of coverage and event-driven wakeup are given in Appendix~\ref{sec:appendix-algorithm-tools}.

\subsection{MLLM-Guided Decisions}

The two members share an MLLM with role-specific prompts. Training on Train Examples updates this model while the perception tools remain frozen. The shared policy also selects movement and observation actions at routine or recovery checks. Four key judgments connect aerial discovery to ground confirmation.

For the UAV, \textbf{Existence Judgment} uses the reference image and current aerial observations to decide whether a suspected target warrants further inspection. \textbf{Instance Judgment} then compares numbered candidate crops with the reference to select a candidate for handoff. Camera geometry projects the selected candidate to a world location, which is sent to the UGV in a text request.

For the UGV, \textbf{Request Acceptance} evaluates an incoming request together with its current observations and search state, deciding whether to accept or defer it. After acceptance, algorithmic routing brings the UGV to the candidate. \textbf{Final Ground Verification} compares the reference with the stationary six-view observations and returns MATCH, NON\_MATCH, or UNKNOWN. A MATCH verdict triggers the terminal confirmation. The other verdicts release the candidate and resume coverage. Aerial search and ground inspection proceed asynchronously, allowing the UAV to continue searching while the UGV approaches or verifies a candidate. The call interface and verdict rules are given in Appendix~\ref{sec:appendix-mllm}.

\begin{table*}[!t]
  \centering
  \fontsize{8}{9.5}\selectfont
  \renewcommand{\arraystretch}{0.9}
  \setlength{\tabcolsep}{3.3pt}
  \caption{Main results on Unseen-Test300 ($n=300$). Each backbone is evaluated with AGOS-Agent and Baseline-Agent. $\Delta$SR is the gain in pp over the paired baseline, or over zero-shot for SFT. SR$_{\mathrm{Easy}}$/SR$_{\mathrm{Mid}}$/SR$_{\mathrm{Hard}}$ are per-difficulty SRs; other metrics are pooled. ``(new)'' marks metrics introduced here. Adapted systems retain their own planners and invocation schedules; AeroDuo-OS additionally uses PilotLLM.}
  \label{tab:main}
  \begin{tabular}{@{}lcccc|ccc|cc|cccc@{}}
    \toprule
    & \multicolumn{4}{c|}{\emph{Navigation}} & \multicolumn{3}{c|}{\emph{Cooperation (new)}} & \multicolumn{2}{c|}{\emph{Efficiency (new)}} & \multicolumn{4}{c@{}}{} \\
    Model / configuration & SR$\uparrow$ & OSR$\uparrow$ & SPL$\uparrow$ & NE$\downarrow$ & ASE$\uparrow$ & ACG$\uparrow$ & GVA$\uparrow$ & DS$\downarrow$ & DSR$\uparrow$ & $\Delta$SR$\uparrow$ & SR$_{\mathrm{Easy}}$$\uparrow$ & SR$_{\mathrm{Mid}}$$\uparrow$ & SR$_{\mathrm{Hard}}$$\uparrow$ \\
    \midrule
    \multicolumn{14}{@{}l}{\emph{On-device MLLMs}} \\
    Qwen3.5-4B (AGOS) & 37.67 & 56.33 & 19.32 & 96.11 & 47.17 & 55.55 & \textbf{92.59} & 36.3 & 2.48 & \textbf{+7.00} & 65.0 & 28.0 & 20.0 \\
    \quad Baseline-Agent & 30.67 & 54.33 & 17.27 & 109.68 & 44.92 & 54.54 & 91.68 & 55.5 & 1.92 &  & 63.0 & 20.0 & 9.0 \\
    \addlinespace[1pt]
    Qwen3-VL-4B (AGOS) & 28.00 & 60.67 & 13.58 & 103.45 & 51.07 & 49.55 & 73.77 & 32.5 & 1.58 & \textbf{+1.67} & 51.0 & 18.0 & 15.0 \\
    \quad Baseline-Agent & 26.33 & 36.67 & 21.83 & 103.44 & 43.97 & 45.12 & 80.77 & 62.7 & 1.20 &  & 48.0 & 16.0 & 15.0 \\
    \addlinespace[1pt]
    Qwen3.5-2B (AGOS) & 18.67 & 45.67 & 11.67 & 125.20 & 46.28 & 48.86 & 87.93 & 55.7 & 1.23 & \textbf{+3.00} & 46.0 & 1.0 & 9.0 \\
    \quad Baseline-Agent & 15.67 & 42.00 & 9.69 & 127.28 & 40.86 & 47.07 & 87.83 & 72.0 & 1.03 &  & 40.0 & 0.0 & 7.0 \\
    \addlinespace[1pt]
    InternVL3\_5-4B (AGOS) & 14.67 & 40.67 & 9.96 & 117.59 & 54.90 & 48.08 & 91.21 & 54.0 & 1.18 & \textbf{+2.33} & 32.0 & 3.0 & 9.0 \\
    \quad Baseline-Agent & 12.33 & 43.33 & 7.96 & 120.42 & 46.44 & 47.92 & 86.90 & 69.2 & 0.82 &  & 33.0 & 2.0 & 2.0 \\
    \addlinespace[1pt]
    Gemma-3-4B (AGOS) & 6.33 & 36.67 & 2.96 & 142.61 & 45.11 & 38.65 & 18.52 & 48.6 & 0.56 & $-$3.33 & 18.0 & 0.0 & 1.0 \\
    \quad Baseline-Agent & 9.67 & 35.67 & 4.42 & 135.63 & 38.86 & 40.23 & 22.83 & 57.8 & 0.77 &  & 28.0 & 1.0 & 0.0 \\
    \addlinespace[1pt]
    InternVL3-2B (AGOS) & 6.33 & 27.33 & 4.72 & 156.27 & 39.74 & 41.60 & 15.83 & 54.5 & 0.78 & \textbf{+0.33} & 17.0 & 1.0 & 1.0 \\
    \quad Baseline-Agent & 6.00 & 27.67 & 4.76 & 157.83 & 36.61 & 40.61 & 15.79 & 62.7 & 0.78 &  & 18.0 & 0.0 & 0.0 \\
    \midrule
    \multicolumn{14}{@{}l}{\emph{API models}} \\
    GPT-6 Astra (AGOS) & 52.33 & 72.00 & 29.35 & 68.16 & 56.66 & \textbf{85.16} & 88.69 & 27.6 & 3.51 & \textbf{+3.33} & 79.0 & 40.0 & \textbf{38.0} \\
    \quad Baseline-Agent & 49.00 & \textbf{74.33} & 28.22 & 71.74 & 52.03 & 79.24 & 89.91 & 53.0 & 2.61 &  & 80.0 & 38.0 & 29.0 \\
    \addlinespace[1pt]
    GLM-5.3 Flash (AGOS) & 44.33 & 69.67 & 22.82 & 82.50 & 53.34 & 71.34 & 86.45 & 29.5 & 3.00 & \textbf{+3.67} & 75.0 & 33.0 & 25.0 \\
    \quad Baseline-Agent & 40.67 & 68.00 & 21.57 & 84.22 & 45.38 & 66.13 & 88.05 & 54.8 & 2.34 &  & 77.0 & 25.0 & 20.0 \\
    \addlinespace[1pt]
    DeepSeek-4.1 Flash (AGOS) & 37.33 & 64.33 & 19.62 & 92.17 & 54.54 & 71.88 & 88.21 & 31.9 & 2.59 & $-$0.67 & 69.0 & 19.0 & 24.0 \\
    \quad Baseline-Agent & 38.00 & 65.00 & 20.71 & 88.30 & 51.16 & 69.84 & 86.97 & 56.5 & 2.16 &  & 70.0 & 24.0 & 20.0 \\
    \midrule
    \multicolumn{14}{@{}l}{\emph{Adapted existing systems}} \\
    AGC-VLN-OS~\cite{zhang2026agcvln} & 31.33 & 42.33 & 17.63 & 109.63 & 28.92 & 51.78 & 33.90 & 36.9 & 1.70 & -- & 68.0 & 15.0 & 11.0 \\
    AeroDuo-OS~\cite{wu2025aeroduo} & 28.67 & 49.67 & 14.51 & 102.90 & \textbf{57.23} & -- & 82.69 & 66.9 & 1.78 & -- & 57.0 & 12.0 & 17.0 \\
    \midrule
    \multicolumn{14}{@{}l}{\emph{Post-trained AGOS-Agent}} \\
    AGOS-Agent + SFT (Ours) & \textbf{57.00} & \textbf{74.33} & \textbf{33.23} & \textbf{65.19} & 52.85 & 81.56 & 78.38 & \textbf{24.2} & \textbf{3.77} & \textbf{+19.33} & \textbf{88.0} & \textbf{46.0} & 37.0 \\
    \bottomrule
  \end{tabular}
\end{table*}

\section{Experiments}
\label{sec:exp}

We evaluate overall task performance across backbones (Table~\ref{tab:main}) and the contribution of MLLM judgments through decision-source controls (Table~\ref{tab:coop2x2}). We then assess fine-tuning on Train Examples (Table~\ref{tab:sft}) and the effects of joint coverage and event-driven wakeup (Table~\ref{tab:ablation}).


\subsection{Experimental Setup}
\label{sec:exp-setup}

On-device models use greedy decoding with a fixed seed and a context length of at most 32{,}768 tokens. Further details of the budget, hardware, and inference stack are given in Appendix~\ref{sec:appendix-settings}. The main experiment, the first auxiliary experiment, and the ablation use Unseen-Test300 (Town03), and the second auxiliary experiment is additionally reported on Seen-Test450.

A supplementary framework comparison uses an MLLM control with basic observation, action, memory, and communication support, including the shared online map and candidate state. Baseline-Agent adds perception tools, dedicated verification calls, and algorithmic routing. AGOS-Agent adds joint coverage and event-driven wakeup. The control is documented as Minimal-Agent in Appendix~\ref{sec:appendix-minimal}.

\subsection{Main Results}
\label{sec:exp-main}

AGOS-Agent improves SR over Baseline-Agent on seven of nine backbones and reduces model calls on all nine (Table~\ref{tab:main}). The SR gain depends on the backbone, while the call reduction holds across both on-device and API models. For Qwen3.5-4B, model calls decrease from 55.5 to 36.3 per episode. The strongest zero-shot model, GPT-6 Astra, achieves an SR of 52.33\%. Adapted AGC-VLN-OS and AeroDuo-OS achieve SRs of 31.33\% and 28.67\%, respectively. They retain their own planners and invocation schedules, including AeroDuo's additional PilotLLM for region guidance (Appendix~\ref{sec:appendix-adapted}).

SR decreases from Easy to Hard as the search region expands and look-alike distractors interfere. Reaching a legal confirmation position also leaves a substantial gap to task completion. Qwen3.5-4B achieves an OSR of 56.33\% and an SR of 37.67\%. Among the 56 episodes that arrive but fail, 53 exhaust the budget and three confirm a wrong vehicle. Budget exhaustion accounts for most failures after legal ground reach (Appendix~\ref{sec:appendix-faildec-abl}).

\textbf{Framework comparison.} With Qwen3.5-4B on Unseen-Test300, the basic-support MLLM control, Baseline-Agent, and AGOS-Agent achieve SRs of 15.33\%, 30.67\%, and 37.67\%, respectively (Table~\ref{tab:minimal}). The largest control-to-Baseline SR gain occurs on Easy, from 21.0\% to 63.0\%. On Hard, their SRs are 12.0\%, 9.0\%, and 20.0\%, respectively.

\begin{table}[!t]
  \centering
  \fontsize{8}{9.5}\selectfont
  \setlength{\tabcolsep}{2.8pt}
  \caption{Hybrid and rule-based configurations on Unseen-Test300 ($n=300$ each). H combines an MLLM with the agent framework; R uses the same framework with similarity-threshold decisions. Per-tier SRs appear in Appendix~\ref{sec:exp-2x2}.}
  \label{tab:coop2x2}
  \begin{tabular}{@{}lllcccccc@{}}
    \toprule
    Configuration & UAV & UGV & SR$\uparrow$ & OSR$\uparrow$ & SPL$\uparrow$ & NE$\downarrow$ & DS$\downarrow$ & DSR$\uparrow$ \\
    \midrule
    \textbf{C1 (Ours)} & H & H & \textbf{37.67} & \textbf{56.33} & \textbf{19.32} & \textbf{96.11} & 36.3 & 2.48 \\
    C2 & H & R & 24.67 & 35.33 & 18.45 & 115.67 & 12.1 & 3.84 \\
    C3 & R & H & 23.33 & 49.00 & 12.54 & 118.53 & \textbf{10.1} & \textbf{5.16} \\
    C4 & R & R & 16.00 & 30.00 & 14.33 & 132.72 & -- & -- \\
    \midrule
    C5 & -- & H & 32.33 & 49.00 & 18.42 & 99.97 & 64.4 & 2.09 \\
    C6 & -- & R & 10.00 & 13.00 & 7.24 & 131.94 & -- & -- \\
    \bottomrule
  \end{tabular}
\end{table}

\subsection{Controlled Comparison of Decision Configurations}

Table~\ref{tab:coop2x2} shows that hybrid decisions on both members give the highest overall SR, while the UGV-only hybrid C5 outperforms the mixed configurations C2 and C3. Using Qwen3.5-4B, C1--C4 compare hybrid (H) and rule-based (R) decision sources within the same AGOS-Agent framework. C5--C6 are UGV-only controls. The fully rule-based C4 achieves an SR of 16.00\%, below the fully hybrid C1. C4 and C6 use rule-based decisions throughout, so DS and DSR are marked ``--''.

The failure profiles reveal different costs of rule-based decisions. Wrong-target confirmation accounts for 93--100\% of failures when ground decisions use similarity-threshold rules. In C3, rule-based aerial nominations accompany budget exhaustion in 89.1\% of failures and an SPL of 12.54. This pattern is consistent with unreliable nominations diverting ground travel toward unproductive regions (Appendix~\ref{sec:appendix-faildec}). C1 has a higher overall SR than C5 while using 36.3 versus 64.4 model calls. The ordering reverses on Hard, where C5 achieves an SR of 33.0\% and C1 achieves 20.0\%. The value of cooperation therefore depends on both decision quality and task difficulty.

\subsection{Fine-Tuning on Train Examples}

\begin{table}[!t]
  \centering
  \fontsize{8}{9.5}\selectfont
  \setlength{\tabcolsep}{2.6pt}
  \caption{Effect of post-training with Train Examples using Qwen3.5-4B, with 100 episodes per difficulty tier on Unseen-Test300 and 150 on Seen-Test450. Per-tier SRs are given in Appendix~\ref{sec:exp-sft}.}
  \label{tab:sft}
  \begin{tabular}{@{}lccccccc@{}}
    \toprule
    Method & SR$\uparrow$ & $\Delta$SR$\uparrow$ & OSR$\uparrow$ & SPL$\uparrow$ & NE$\downarrow$ & DS$\downarrow$ & DSR$\uparrow$ \\
    \midrule
    \multicolumn{8}{@{}l}{\emph{Unseen-Test300}} \\
    Zero-shot & 37.67 & & 56.33 & 19.32 & 96.11 & 36.3 & 2.48 \\
    Joint training (Ours) & \textbf{57.00} & \textbf{+19.33} & \textbf{74.33} & \textbf{33.23} & \textbf{65.19} & 24.2 & 3.77 \\
    Role-split training & 56.67 & +19.00 & 73.33 & 32.19 & 65.33 & \textbf{23.6} & \textbf{3.83} \\
    \midrule
    \multicolumn{8}{@{}l}{\emph{Seen-Test450}} \\
    Zero-shot & 28.22 & & 64.89 & 12.34 & 113.83 & 39.8 & 2.03 \\
    Joint training (Ours) & \textbf{52.00} & \textbf{+23.78} & \textbf{74.22} & \textbf{23.18} & \textbf{74.42} & 25.9 & \textbf{3.48} \\
    Role-split training & 50.00 & +21.78 & 71.11 & 22.11 & 77.12 & \textbf{25.1} & 3.46 \\
    \bottomrule
  \end{tabular}
\end{table}

We evaluate two strategies for fine-tuning the backbone on Train Examples: joint training and role-split training. Joint training fits a shared Qwen3.5-4B LoRA adapter to both roles using role-specific prompts, whereas role-split training fits a separate adapter for each role. Both use the same 6{,}014 filtered Train Examples and retain the AGOS-Agent inference procedure. Table~\ref{tab:sft} shows higher SR on both test sets with either strategy. Joint and role-split training differ in SR by at most 2.00 pp, supporting a shared adapter as a compact deployment choice.

Fine-tuning improves aerial candidate guidance and legal ground reach on Unseen-Test300. In Table~\ref{tab:main}, ACG increases from 55.55 to 81.56, OSR rises by 18.00 pp, and model calls decrease from 36.3 to 24.2 per episode. GVA falls from 92.59\% to 78.38\%. This event-pooled score reflects both the cases encountered and the policy's verdicts. Greater exposure to Hard confirmation cases and a stronger tendency to confirm are plausible contributors to the decline. Mistaken approaches remain sparsely represented in the Train Examples, motivating broader supervision of these cases (Appendix~\ref{sec:exp-sft}).

\begin{table}[!t]
  \centering
  \fontsize{8}{9.5}\selectfont
  \setlength{\tabcolsep}{2.8pt}
  \caption{A 2$\times$2 ablation of joint coverage (JC) and event-driven wakeup (EW) on Unseen-Test300 ($n=300$ per variant). Brackets give episode-paired bootstrap 95\% CIs; paired McNemar tests with Holm correction are in Appendix~\ref{sec:appendix-stats}. EW uses event-gated wakeup when enabled and decision-point wakeup when disabled.}
  \label{tab:ablation}
  \begin{tabular}{@{}lcccccc@{}}
    \toprule
    Variant & SR$\uparrow$ & $\Delta$SR [95\% CI]$\uparrow$ & OSR$\uparrow$ & SPL$\uparrow$ & DS$\downarrow$ & DSR$\uparrow$ \\
    \midrule
    Baseline  &  30.67  &  --  &  54.33  & 17.27 & 55.5 &  1.92  \\
    w/o EW  &  29.67  &  $-1.00$ [$-5.33$, $+3.33$]  &  50.00  & 17.02 & 56.3 &  1.96  \\
    w/o JC  &  34.33  &  $+3.67$ [$-1.00$, $+8.33$]  &  \textbf{58.00}  & 19.14 & \textbf{35.4} &  2.43  \\
    Full (Ours)  &  \textbf{37.67}  &  \textbf{+7.00} [$+2.00$, $+12.00$]  &  56.33  & \textbf{19.32} & 36.3 &  \textbf{2.48}  \\
    \bottomrule
  \end{tabular}
\end{table}

\subsection{Ablation Study}
\label{sec:exp-abl}

Combining joint coverage and event-driven wakeup gives the highest SR in the 2$\times$2 ablation (Table~\ref{tab:ablation}). Full reaches 37.67\%, compared with 30.67\% for Baseline-Agent. Enabling only EW yields 34.33\%, whereas enabling only JC yields 29.67\%.

Event-driven wakeup reduces model calls under both coverage settings: from 55.5 to 35.4 with fixed coverage and from 56.3 to 36.3 with joint coverage. Wrong-target confirmations range from 22 to 28 episodes across variants, while budget exhaustion ranges from 162 to 187. These results link selective invocation to lower decision cost, with failure differences concentrated in budget exhaustion. Complete failure counts and paired tests appear in Appendices~\ref{sec:appendix-faildec-abl} and~\ref{sec:appendix-stats}.

\section{Conclusion and Limitations}
\label{sec:concl}

AGOS connects urban exploration with ground-level instance confirmation, making coordinated discovery and verification central to embodied air-ground search. AGOS-Bench defines the task and evaluation, AGOS-Dataset supplies Expert Trajectories and Train Examples, and AGOS-Agent combines algorithmic search with trainable MLLM decisions. Across nine backbones, AGOS-Agent improves SR on seven and reduces model calls on all nine. Fine-tuning Qwen3.5-4B raises unseen SR from 37.67\% to 57.00\%. Together, these resources support research on how aerial discoveries become reliable ground actions, with shared task interfaces for comparing planning, cooperation, and learned verification.

The current setting uses parked vehicles, ideal communication, and discrete actions. Further development should strengthen instruction following and structured output, broaden supervision for mistaken approaches through online iterative training, and extend evaluation toward physical systems. Dynamic traffic, communication delays, and GPS drift would test the robustness of candidate handoff and ground verification under conditions closer to deployment.

{\footnotesize
\bibliographystyle{ieeenat_fullname}
\bibliography{refs}
}

\clearpage
\appendix
\raggedbottom
\twocolumn[{
\begin{@twocolumnfalse}
\begin{center}
{\Large\bfseries Appendix}
\end{center}
\vspace*{4pt}
\centering
\captionsetup{aboveskip=5pt,belowskip=3pt}
\includegraphics[width=0.85\textwidth]{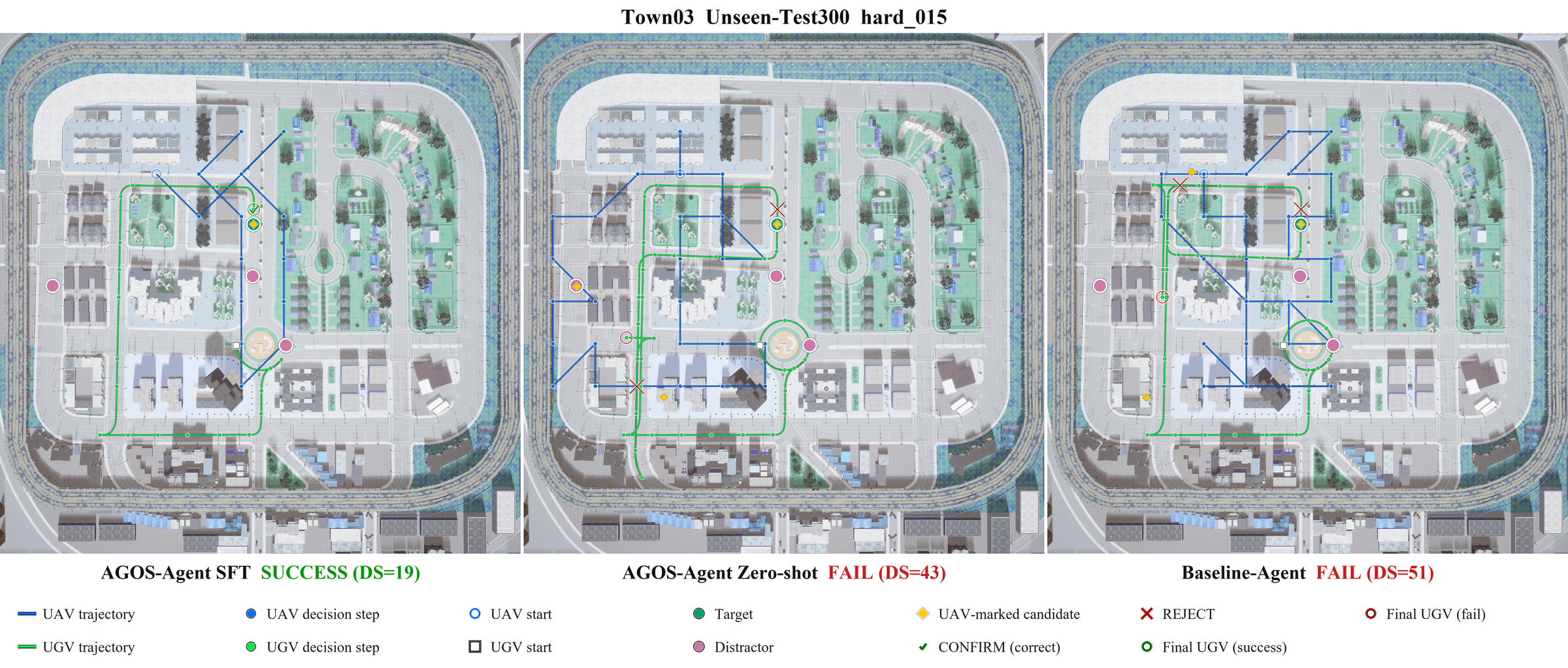}
\captionof{figure}{Example A on Unseen-Test300 (Town03, Hard). The three trajectories show how the ground verdict on the first candidate affects the rest of the search. DS counts MLLM decision calls by both roles.}
\label{fig:demo-a}
\par\vspace{2pt}
\includegraphics[width=0.85\textwidth]{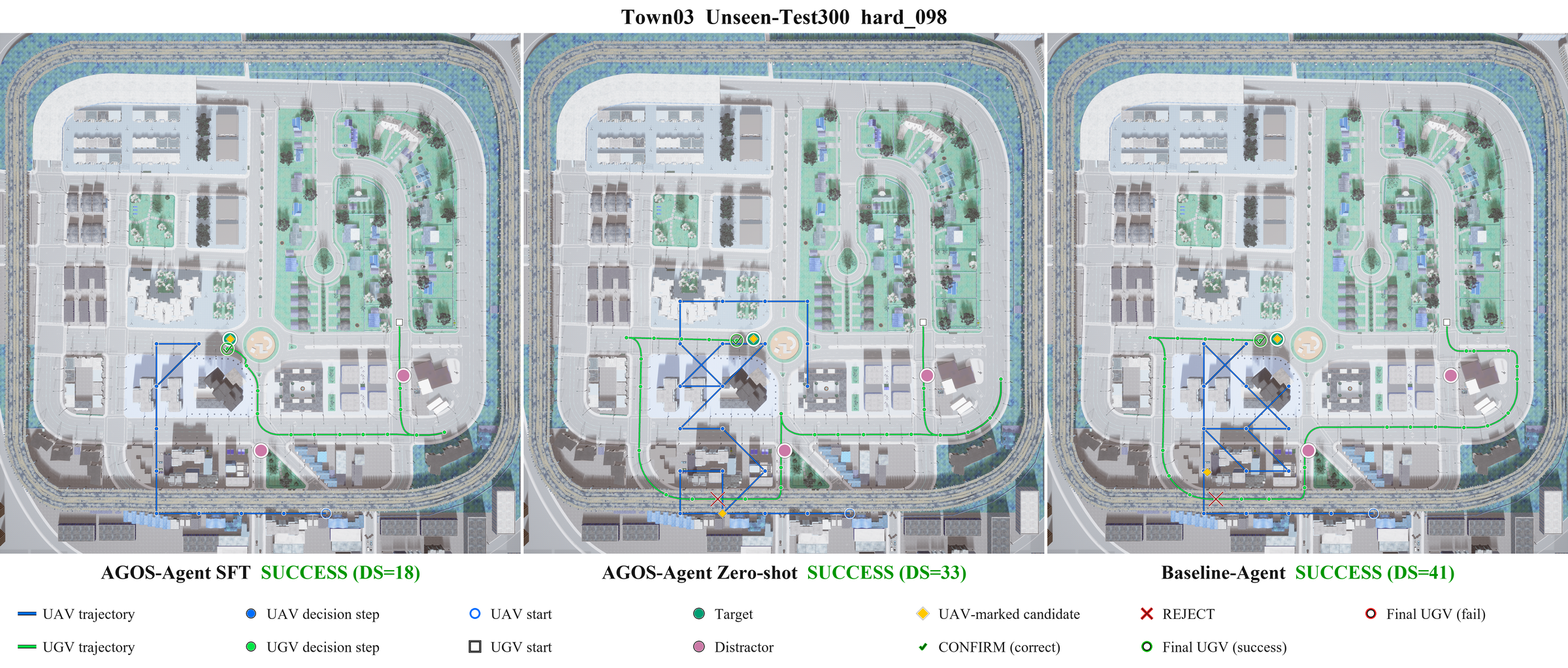}
\captionof{figure}{Example B on Unseen-Test300 (Town03, Hard). All three configurations complete the task. Only AGOS-Agent + SFT confirms its first candidate.}
\label{fig:demo-b}
\end{@twocolumnfalse}
}]

\section{Qualitative Examples}
\label{sec:appendix-demo}

\subsection{Example A: Early Rejection of a True Target}
\label{sec:appendix-demo-a}

In Fig.~\ref{fig:demo-a}, all three configurations nominate the same first candidate within 14\,m of the target at task steps 3--7. AGOS-Agent + SFT identifies the candidate as the reference vehicle (MATCH) and succeeds in 37 task steps (DS 19). Both zero-shot AGOS-Agent and Baseline-Agent judge it to be a different vehicle (NON\_MATCH) and resume coverage outside the rejected region. After 80 task steps, they finish 177.8 and 147.5\,m from the target, respectively (DS 43 and 51).

\subsection{Example B: Candidate Selection and Decision Cost}
\label{sec:appendix-demo-b}

In Fig.~\ref{fig:demo-b}, AGOS-Agent + SFT confirms its first candidate at task step 27, 9.5\,m from the target (DS 18). The first candidates inspected by zero-shot AGOS-Agent and Baseline-Agent are look-alikes 154.5 and 161.5\,m from the target, respectively. Both reject the look-alike and later confirm from the same pose, 15.9\,m from the target. Confirmation occurs at steps 54 and 48 (DS 33 and 41).

\twocolumn[{
\begin{@twocolumnfalse}
\section{Additional Experiments}
\label{sec:appendix-experiments}

\subsection{Framework Comparison}
\label{sec:appendix-minimal}
\par\smallskip
\noindent\begin{minipage}{\textwidth}
\centering
  \fontsize{8.5}{10}\selectfont
  \renewcommand{\arraystretch}{1.0}
  \setlength{\tabcolsep}{2.0pt}
  \captionof{table}{Framework comparison on Unseen-Test300 ($n=300$, 100 episodes per tier). NE is in meters and DS in calls per episode. Other scores are percentages. The other two configurations use the runs reported in the main paper.}
  \label{tab:minimal}
  \begin{tabular}{@{}lcccccccccccc@{}}
    \toprule
    Configuration & SR$\uparrow$ & OSR$\uparrow$ & SPL$\uparrow$ & NE$\downarrow$ & ASE$\uparrow$ & ACG$\uparrow$ & GVA$\uparrow$ & DS$\downarrow$ & DSR$\uparrow$ & SR$_{\mathrm{Easy}}$$\uparrow$ & SR$_{\mathrm{Mid}}$$\uparrow$ & SR$_{\mathrm{Hard}}$$\uparrow$ \\
    \midrule
    Minimal-Agent & 15.33 & 42.33 & 7.41 & 137.76 & 47.33 & 38.53 & 64.79 & 70.0 & 1.01 & 21.0 & 13.0 & 12.0 \\
    Baseline-Agent & 30.67 & 54.33 & 17.27 & 109.68 & 44.92 & 54.54 & 91.68 & 55.5 & 1.92 & 63.0 & 20.0 & 9.0 \\
    AGOS-Agent & 37.67 & 56.33 & 19.32 & 96.11 & 47.17 & 55.55 & 92.59 & 36.3 & 2.48 & 65.0 & 28.0 & 20.0 \\
    \bottomrule
  \end{tabular}
\end{minipage}\par
\par\smallskip
\noindent\begin{minipage}{\textwidth}
\centering
  \small
  \setlength{\tabcolsep}{4pt}
  \captionof{table}{Outcome decomposition on Unseen-Test300 (episodes, $n=300$ per configuration). Wrong-target denotes a terminal CONFIRM at a pose that fails the oracle confirmation criteria. Budget denotes step-budget exhaustion. Patrol-loop and Nav-stall are method-side failures.}
  \label{tab:minimal-failures}
  \begin{tabular}{@{}lccccc@{}}
    \toprule
    Configuration & Success & Wrong-target & Budget & Patrol-loop & Nav-stall \\
    \midrule
    Minimal-Agent & 46 & 25 & 140 & 81 & 8 \\
    Baseline-Agent & 92 & 28 & 180 & 0 & 0 \\
    AGOS-Agent & 113 & 25 & 162 & 0 & 0 \\
    \bottomrule
  \end{tabular}
\end{minipage}\par
\par\smallskip
\noindent\begin{minipage}{\textwidth}
\centering
  \fontsize{9}{10.5}\selectfont
  \setlength{\tabcolsep}{2.6pt}
  \captionof{table}{Paired success rates on Unseen-Test300 ($n=300$). The two discordant counts give successes unique to the first and second configurations, respectively. $\Delta$SR is the second SR minus the first, in percentage points (pp). The intervals are unadjusted 95\% episode-paired percentile intervals (10,000 resamples, seed 20261110). Two-sided exact McNemar tests use Holm correction for these three exploratory comparisons.}
  \label{tab:minimal-paired}
  \begin{tabular}{@{}lcccccc@{}}
    \toprule
    Comparison & Former-only & Latter-only & $\Delta$SR (pp) & 95\% CI (pp) & McNemar $p$ & Holm(3) $p$ \\
    \midrule
    Minimal-Agent $\rightarrow$ Baseline-Agent & 27 & 73 & +15.33 & [9.00, 21.67] & $4.69\times10^{-6}$ & $9.38\times10^{-6}$ \\
    Baseline-Agent $\rightarrow$ AGOS-Agent & 19 & 40 & +7.00 & [2.00, 12.00] & $8.64\times10^{-3}$ & $8.64\times10^{-3}$ \\
    Minimal-Agent $\rightarrow$ AGOS-Agent & 24 & 91 & +22.33 & [16.00, 28.67] & $2.26\times10^{-10}$ & $6.77\times10^{-10}$ \\
    \bottomrule
  \end{tabular}
\end{minipage}\par
\end{@twocolumnfalse}
}]

The three configurations share a candidate ledger. Their shared online map is the town's waypoint and road-network graph, fixed at episode-set construction and shared by the UAV and UGV (Section~\ref{sec:appendix-waypoints}). It defines legal moves and the coordinate frame for candidate records. The model receives no map image. Minimal-Agent selects legal actions from camera views, the target reference, candidate records in text and structured form, and messages. The UAV also receives a position-free image of the UGV for teammate recognition. Baseline-Agent adds perception tools, dedicated verification calls, and algorithmic routing. AGOS-Agent further adds joint coverage and event-driven wakeup.

With Qwen3.5-4B, a common budget, and the same success criteria (Table~\ref{tab:minimal}), SR, SPL, ACG, and DSR rise from Minimal-Agent through Baseline-Agent to AGOS-Agent, while NE and DS fall. Minimal-Agent has the highest ASE (47.33\%). The three configurations exhaust the budget in 140, 180, and 162 episodes, respectively. All 81 patrol-loop failures and eight navigation stalls occur with Minimal-Agent (Table~\ref{tab:minimal-failures}).

Table~\ref{tab:minimal-paired} compares complete configurations. Each solves episodes its comparator misses. On Hard, Minimal-Agent alone solves 11 episodes and Baseline-Agent alone solves eight in their paired comparison. Their SRs (12\% and 9\%) reverse the overall ordering.

\subsection{Adapted Cooperative Systems}
\label{sec:appendix-adapted}

We adapt AGC-VLN and AeroDuo to visual-reference object search~\cite{zhang2026agcvln,wu2025aeroduo}. Both use Qwen3.5-4B for decision calls, an online visual map built from UAV observations, and the benchmark's discrete actions and ground-confirmation criteria. Each retains its own planning and model-call schedule (settings in Appendix~\ref{sec:appendix-adapted-settings}).

AGC-VLN-OS localizes candidates in aerial images and maps ten predicted image waypoints to an executable road route. The UGV verifies candidates in its six views. AeroDuo-OS uses the released PilotLLM (Qwen2-VL-2B with its adapter and probability head) to predict search regions. The UGV searches these regions with aerial guidance, road-graph routing, Grounding DINO proposals, and six-view confirmation. Region-level predictions do not supply the individual candidates required for ACG.

\subsection{Decision-Source Controls}
\label{sec:exp-2x2}

Configurations C1--C4 combine hybrid (H) and rule-based (R) decisions for the UAV and UGV. C5 and C6 use only the UGV. All six share the Unseen-Test300 tasks, budget, and success criteria (Table~\ref{tab:coop2x2full}).

\twocolumn[{
\begin{@twocolumnfalse}

\noindent\begin{minipage}{\textwidth}
\centering
  \small
  \setlength{\tabcolsep}{2.5pt}
  \captionof{table}{Decision-source controls on Unseen-Test300 ($n=300$ each). H combines an MLLM with the agent framework. R uses the same framework with similarity-threshold decisions. A dash marks an absent member or an inapplicable metric. Bold marks the best value in each column.}
  \label{tab:coop2x2full}
  \begin{tabular}{@{}lllccccccccc@{}}
    \toprule
    Configuration & UAV & UGV & SR$\uparrow$ & OSR$\uparrow$ & SPL$\uparrow$ & NE$\downarrow$ & DS$\downarrow$ & DSR$\uparrow$ & SR$_{\mathrm{Easy}}$$\uparrow$ & SR$_{\mathrm{Mid}}$$\uparrow$ & SR$_{\mathrm{Hard}}$$\uparrow$ \\
    \midrule
    \textbf{C1 (Ours)}  &  H  &  H  &  \textbf{37.67}  &  \textbf{56.33}  & \textbf{19.32} &  \textbf{96.11}  &  36.3  &  2.48  &  \textbf{65.0}  &  \textbf{28.0}  &  20.0  \\
    C2  &  H  &  R  &  24.67  &  35.33  & 18.45 &  115.67  &  12.1  &  3.84  &  58.0  &  10.0  &  6.0  \\
    C3  &  R  &  H  &  23.33  &  49.00  & 12.54 &  118.53  &  \textbf{10.1}  &  \textbf{5.16}  &  54.0  &  8.0  &  8.0  \\
    C4  &  R  &  R  &  16.00  &  30.00  & 14.33 &  132.72  &  --  &  --  &  46.0  &  1.0  &  1.0  \\
    \midrule
    C5  &  --  &  H  &  32.33  &  49.00  & 18.42 &  99.97  &  64.4  &  2.09  &  44.0  &  20.0  &  \textbf{33.0}  \\
    C6  &  --  &  R  &  10.00  &  13.00  & 7.24 &  131.94  &  --  &  --  &  29.0  &  0.0  &  1.0  \\
    \bottomrule
  \end{tabular}
\end{minipage}\par
\par\smallskip
\noindent\begin{minipage}{\textwidth}
\centering
  \small
  \setlength{\tabcolsep}{4pt}
  \captionof{table}{Failure-mode decomposition on Unseen-Test300 (episodes, $n=300$ per configuration). Wrong-target counts terminal CONFIRM actions at poses that fail the oracle criteria. Budget counts step-budget exhaustion and Stall counts method-side stalls. The OSR column counts episodes that reach a legal confirmation position. H $=$ hybrid configuration and R $=$ rule-based configuration, both using the agent framework.}
  \label{tab:faildec}
  \begin{tabular}{@{}lccccc@{}}
    \toprule
    Configuration & Success & Wrong-target & Budget & Stall & OSR \\
    \midrule
    \multicolumn{6}{@{}l}{\emph{Decision sources (C1--C6)}} \\
    C1 (H+H) & 113 & 25 & 162 & 0 & 169 \\
    C2 (H+R) & 74 & 211 & 15 & 0 & 106 \\
    C3 (R+H) & 70 & 25 & 205 & 0 & 147 \\
    C4 (R+R) & 48 & 247 & 5 & 0 & 90 \\
    C5 (--,H) & 97 & 61 & 129 & 13 & 147 \\
    C6 (--,R) & 30 & 270 & 0 & 0 & 39 \\
    \bottomrule
  \end{tabular}
\end{minipage}\par
\par\smallskip
\noindent\begin{minipage}{\textwidth}
\centering
  \scriptsize
  \setlength{\tabcolsep}{3pt}
  \captionof{table}{Operating conditions for the decision-source controls. The rule similarity threshold $\tau$ applies to nomination and to confirmation by a rule-based UGV. The shared candidate ledger permits one active candidate. Ordinary suppression of a rejected region also applies when rejection is available and the extra cooldown is off.}
  \label{tab:decision-conditions}
  \begin{tabular}{@{}llc p{0.16\textwidth} p{0.24\textwidth} p{0.25\textwidth}@{}}
    \toprule
    Configuration & UAV/UGV & Rule similarity & UGV verdicts & Candidate nomination & Rejection-region / reachability condition \\
    \midrule
    C1 & H/H & -- & CONFIRM and REJECT & Hybrid UAV and UGV; one active candidate & No rejection-region cooldown \\
    C2 & H/R & UGV $\tau=0.55$ & CONFIRM only & Hybrid UAV and rule UGV; one active candidate & No rejection-region cooldown; confirmation budget 1 \\
    C3 & R/H & UAV $\tau=0.55$ & CONFIRM and REJECT & Rule UAV; UGV can nominate when no candidate is active & Cooldown 12 steps, one reopen; event-driven UGV consultation limits self-nomination reachability \\
    C4 & R/R & UAV and UGV $\tau=0.55$ & CONFIRM only & Rule UAV and UGV; one active candidate & No rejection-region cooldown; confirmation budget 1 \\
    C5 & --/H & -- & CONFIRM and REJECT & UGV self-nomination is reachable at stepwise decisions & Cooldown 12 steps, one reopen \\
    C6 & --/R & UGV $\tau=0.55$ & CONFIRM only & Rule UGV nomination & No rejection-region cooldown; confirmation budget 1 \\
    \bottomrule
  \end{tabular}
\end{minipage}\par
\end{@twocolumnfalse}
}]

C1 leads C5 in overall SR (37.67\% vs.\ 32.33\%). On Hard, the UGV-only C5 reaches 33\%, compared with 20\% for C1. In the larger search regions, following a handoff to a look-alike may add ground travel.

\subsubsection{Failure-Mode Decomposition}
\label{sec:appendix-faildec}

Table~\ref{tab:faildec} decomposes the outcomes of the six configurations in Table~\ref{tab:coop2x2full}.

Wrong-target confirmation accounts for 93.4--100\% of failures in configurations with a rule-based ground member (C2, C4, C6), compared with 10.9--30.0\% for a hybrid ground member (C1, C3, C5). In C3, which combines rule-based aerial nominations with hybrid ground decisions, 89.1\% of failures are budget exhaustion. This pattern is consistent with unreliable aerial nominations diverting the UGV toward unproductive regions.

Table~\ref{tab:decision-conditions} gives the actions available to each configuration. A rule-based UGV can confirm a candidate and never issues REJECT. C3 and C5 also apply a 12-step cooldown and allow one reopening. These conditions affect which candidates each configuration can inspect.

\twocolumn[{
\begin{@twocolumnfalse}
\subsection{Fine-Tuning Results}
\label{sec:exp-sft}
\par\smallskip
\noindent\begin{minipage}{\textwidth}
\centering
  \small
  \setlength{\tabcolsep}{3pt}
  \captionof{table}{Qwen3.5-4B results on Unseen-Test300 (100 episodes per tier) and Seen-Test450 (150 per tier). $\Delta$SR is the paired difference from zero-shot in percentage points, NE is in meters, and DS is in calls per episode. All other scores are percentages. GVA pools verification events (Table~\ref{tab:gva-support}). The remaining metrics summarize episodes. Bold marks the best value within each panel.}
  \label{tab:sftfull}
  \resizebox{\textwidth}{!}{%
  \begin{tabular}{@{}lccccccccccccc@{}}
    \toprule
    Method & SR$\uparrow$ & $\Delta$SR$\uparrow$ & OSR$\uparrow$ & SPL$\uparrow$ & NE$\downarrow$ & ASE$\uparrow$ & ACG$\uparrow$ & GVA$\uparrow$ & DS$\downarrow$ & DSR$\uparrow$ & SR$_{\mathrm{Easy}}$$\uparrow$ & SR$_{\mathrm{Mid}}$$\uparrow$ & SR$_{\mathrm{Hard}}$$\uparrow$ \\
    \midrule
    \multicolumn{14}{@{}l}{\emph{Unseen-Test300}} \\
    Zero-shot  &  37.67  &   &  56.33  & 19.32 &  96.11  & 47.17 & 55.55 & \textbf{92.59} & 36.3  &  2.48  &  65.0  &  28.0  &  20.0  \\
    Joint training (Ours)  &  \textbf{57.00}  &  \textbf{+19.33}  &  \textbf{74.33}  & \textbf{33.23} &  \textbf{65.19}  & \textbf{52.85} & 81.56 & 78.38 & 24.2  &  3.77  &  \textbf{88.0}  &  \textbf{46.0}  &  \textbf{37.0}  \\
    Role-split training  &  56.67  &  +19.00  &  73.33  & 32.19 &  65.33  & 52.51 & \textbf{81.70} & 76.23 & \textbf{23.6}  &  \textbf{3.83}  &  87.0  &  \textbf{46.0}  &  \textbf{37.0}  \\
    \midrule
    \multicolumn{14}{@{}l}{\emph{Seen-Test450}} \\
    Zero-shot  &  28.22  &   &  64.89  & 12.34 &  113.83  & 44.65 & 50.76 & \textbf{89.44} & 39.8  &  2.03  &  46.7  &  20.7  &  17.3  \\
    Joint training (Ours)  &  \textbf{52.00}  &  \textbf{+23.78}  &  \textbf{74.22}  & \textbf{23.18} &  \textbf{74.42}  & \textbf{50.90} & \textbf{69.14} & 74.76 & 25.9  &  \textbf{3.48}  &  \textbf{64.7}  &  \textbf{46.0}  &  \textbf{45.3}  \\
    Role-split training  &  50.00  &  +21.78  &  71.11  & 22.11 &  77.12  & 49.86 & 67.91 & 69.23 & \textbf{25.1}  &  3.46  &  64.0  &  40.7  &  \textbf{45.3}  \\
    \bottomrule
  \end{tabular}%
  }
\end{minipage}\par
\end{@twocolumnfalse}
}]

Joint training uses one Qwen3.5-4B LoRA adapter for both roles, with role conditioning supplied through prompts. Role-split training uses one adapter per role. Both use the same 6{,}014 filtered Train Examples and retain the AGOS-Agent inference procedure. Full results appear in Table~\ref{tab:sftfull}, and the training parameters are given in Appendix~\ref{sec:appendix-train}.

\textbf{Training strategy.} In these single-seed runs, the two training strategies differ in SR by at most 2.00 pp. We use joint training for the main configuration because it serves both roles with one adapter.

\textbf{Training supervision.} The expert demonstrations cover states encountered by the collection policy. The fine-tuning input contains no NON\_MATCH verdicts or REJECT actions.

\textbf{Ground verification.} On Unseen-Test300, correct rejections account for 412 of 567 verification events with zero-shot AGOS-Agent, three of 222 with joint training, and zero of 223 with role-split training. The policies encounter different sets of events (Table~\ref{tab:gva-support}).

\subsection{Component Ablations}
\label{sec:appendix-ablation}

\subsubsection{Failure-Mode Decomposition}
\label{sec:appendix-faildec-abl}

Table~\ref{tab:faildec-abl} gives the failure counts for the four ablation variants in Table~\ref{tab:ablation} of the main paper.

\noindent\begin{minipage}{\columnwidth}
\centering
  \small
  \setlength{\tabcolsep}{4pt}
  \captionof{table}{Failure-mode decomposition of the ablation variants on Unseen-Test300 (episodes, $n=300$ per variant). Columns follow Table~\ref{tab:faildec}. Baseline reuses the main-experiment baseline runs, and Full reuses the C1 runs.}
  \label{tab:faildec-abl}
  \begin{tabular}{@{}lccccc@{}}
    \toprule
    Variant & Success & Wrong-target & Budget & Stall & OSR \\
    \midrule
    Baseline & 92 & 28 & 180 & 0 & 163 \\
    w/o EW & 89 & 24 & 187 & 0 & 150 \\
    w/o JC & 103 & 22 & 175 & 0 & 174 \\
    Full & 113 & 25 & 162 & 0 & 169 \\
    \bottomrule
  \end{tabular}
\end{minipage}\par

With Full, 169 episodes reach a legal confirmation position and 113 succeed. Of the remaining 56, 53 exhaust the budget and three end in wrong-target confirmation. Another 22 episodes end in wrong-target confirmation without reaching a legal confirmation position. With Baseline-Agent, 68 of the 71 episodes in the OSR--SR gap end in budget exhaustion.

\subsubsection{Statistical Comparisons}
\label{sec:appendix-stats}

The ablation compares joint coverage and event-driven wakeup in a Baseline-anchored $2\times2$ design. Table~\ref{tab:ablation} in the main paper reports unadjusted 95\% CIs from 10,000 episode-paired bootstrap resamples (seed 20261110).

Two-sided exact paired McNemar tests of each variant against Baseline-Agent give $p$-values of 0.7608 (w/o EW), 0.177 (w/o JC), and 0.008641 (Full). The (variant-only, Baseline-only) discordant-pair counts are (20, 23), (33, 22), and (40, 19), respectively; the corresponding Holm-corrected values are 0.7608, 0.354, and 0.0259.

The Baseline-Agent and Full results come from the same runs as the Qwen3.5-4B backbone comparison in Table~\ref{tab:backbone-paired}.

\subsection{Paired Statistical Comparisons}
\label{sec:appendix-paired-stats}

We conducted post hoc analyses of paired outcomes on the same episodes. We report unadjusted 95\% percentile intervals from 10,000 episode-paired bootstrap resamples (seed 20261110). Two-sided exact McNemar tests use the discordant episode pairs. We applied Holm correction across the nine backbones and, separately, across the three training comparisons within each test set. Each SFT contrast compares the two deployed configurations on matched tasks.

\twocolumn[{
\begin{@twocolumnfalse}
\noindent\begin{minipage}{\textwidth}
\centering
  \small
  \setlength{\tabcolsep}{3pt}
  \captionof{table}{Paired backbone comparisons on Unseen-Test300 ($n=300$). The first two counts are episodes solved only by Baseline-Agent or only by AGOS-Agent. $\Delta$SR is AGOS-Agent minus Baseline-Agent in percentage points. The last two columns give the exact and Holm-adjusted $p$-values.}
  \label{tab:backbone-paired}
  \resizebox{\textwidth}{!}{%
  \begin{tabular}{@{}lcccccc@{}}
    \toprule
    Backbone (Baseline-Agent $\rightarrow$ AGOS-Agent) & Former-only & Latter-only & $\Delta$SR (pp) & 95\% CI (pp) & McNemar $p$ & Holm(9) $p$ \\
    \midrule
    Qwen3.5-4B & 19 & 40 & +7.00 & [2.00, 12.00] & $8.641\times10^{-3}$ & $7.777\times10^{-2}$ \\
    Qwen3-VL-4B & 27 & 32 & +1.67 & [-3.33, 6.67] & 0.6029 & 1.0000 \\
    Qwen3.5-2B & 16 & 25 & +3.00 & [-1.00, 7.00] & 0.2110 & 1.0000 \\
    InternVL3\_5-4B & 16 & 23 & +2.33 & [-1.67, 6.33] & 0.3368 & 1.0000 \\
    Gemma-3-4B & 20 & 10 & -3.33 & [-7.00, 0.33] & 0.09874 & 0.7899 \\
    InternVL3-2B & 3 & 4 & +0.33 & [-1.33, 2.00] & 1.0000 & 1.0000 \\
    GPT-6 Astra & 33 & 43 & +3.33 & [-2.00, 9.00] & 0.3019 & 1.0000 \\
    GLM-5.3 Flash & 38 & 49 & +3.67 & [-2.33, 10.00] & 0.2836 & 1.0000 \\
    DeepSeek-4.1 Flash & 47 & 45 & -0.67 & [-7.00, 5.67] & 0.9170 & 1.0000 \\
    \bottomrule
  \end{tabular}%
  }
\end{minipage}\par
\par\smallskip
\noindent\begin{minipage}{\textwidth}
\centering
  \small
  \setlength{\tabcolsep}{3pt}
  \captionof{table}{Paired deployment comparisons. Each row gives successes unique to the first or second configuration and their SR difference (second minus first). Holm correction covers the three comparisons within each test set.}
  \label{tab:sft-paired}
  \resizebox{\textwidth}{!}{%
  \begin{tabular}{@{}lcccccc@{}}
    \toprule
    Comparison & Former-only & Latter-only & $\Delta$SR (pp) & 95\% CI (pp) & McNemar $p$ & Holm(3) $p$ \\
    \midrule
    \multicolumn{7}{@{}l}{\emph{Unseen-Test300 ($n=300$ paired episodes)}} \\
    Zero-shot $\rightarrow$ Joint training & 25 & 83 & +19.33 & [12.67, 25.67] & $1.8829\times10^{-8}$ & $5.6488\times10^{-8}$ \\
    Zero-shot $\rightarrow$ Role-split training & 27 & 84 & +19.00 & [12.33, 25.67] & $5.4620\times10^{-8}$ & $1.0924\times10^{-7}$ \\
    Joint training $\rightarrow$ Role-split training & 12 & 11 & -0.33 & [-3.33, 2.67] & 1.0000 & 1.0000 \\
    \midrule
    \multicolumn{7}{@{}l}{\emph{Seen-Test450 ($n=450$ paired episodes)}} \\
    Zero-shot $\rightarrow$ Joint training & 31 & 138 & +23.78 & [18.67, 29.11] & $2.5736\times10^{-17}$ & $7.7208\times10^{-17}$ \\
    Zero-shot $\rightarrow$ Role-split training & 33 & 131 & +21.78 & [16.44, 27.11] & $5.0755\times10^{-15}$ & $1.0151\times10^{-14}$ \\
    Joint training $\rightarrow$ Role-split training & 24 & 15 & -2.00 & [-4.67, 0.67] & 0.1996 & 0.1996 \\
    \bottomrule
  \end{tabular}%
  }
\end{minipage}\par
\end{@twocolumnfalse}
}]

Table~\ref{tab:backbone-paired} shows a higher observed SR for AGOS-Agent on seven of nine backbones. For Qwen3.5-4B, the difference is +7.00 pp (unadjusted 95\% CI, 2.00 to 12.00). The smallest Holm-adjusted $p$ is 0.07777, above the 0.05 threshold for this family.

In Table~\ref{tab:sft-paired}, both fine-tuned configurations have higher SR than zero-shot AGOS-Agent on both test sets. Relative to joint training, role-split training changes SR by $-0.33$ pp on Unseen-Test300 and $-2.00$ pp on Seen-Test450. These paired comparisons leave the question of equivalence open.

\section{Benchmark Specifications and Metrics}
\label{sec:appendix-benchmark}

\subsection{Observations and Actions}
\label{sec:appendix-actions}

\textbf{UAV.} The UAV flies at 80\,m with a $90^{\circ}$ field of view and a fixed heading. It observes a top-down RGB image at $1024\times576$. Coverage planning uses an effective sampling radius of 25\,m. Its executable actions include moving between grid waypoints in eight horizontal directions, annotating candidates, requesting assistance, and pausing to observe.

\textbf{UGV.} The UGV observes a six-view RGB set ($480\times270$ per view), tiled into a $3\times2$ collage at $1440\times540$. The azimuth angles are $-45^{\circ}$, $0^{\circ}$, $45^{\circ}$, $-135^{\circ}$, $180^{\circ}$, and $135^{\circ}$, with a $90^{\circ}$ field of view. Its executable actions include moving forward along roads, reversing, turning at intersections, confirming, rejecting, and stopping to observe. All actions are discrete and atomic, and movement cost is measured by actual path length. Virtual time advances with action duration, with annotation, confirmation, observation, and turning timed separately. The two members therefore decide asynchronously.

\textbf{Reference image protocol.} The reference image is a single background-free photo taken at a distance of about 12\,m. Its elevation angle is sampled randomly between $10^{\circ}$ and $75^{\circ}$. The capture location differs from the target deployment location, so the reference image does not reveal the deployment position.

\textbf{Action timing.} Annotation, confirmation, and rejection each take 5\,s. STOP\_OBSERVE and WAIT each take 10\,s. Both members move at 5\,m/s, with an additional 1\,s for observation and a minimum movement time of 1\,s. A turn takes 2\,s and a U-turn 4\,s. Travel uses straight-line distance for the UAV and road distance for the UGV. The coverage planner estimates travel time at 10\,m/s for the UAV and 6\,m/s for the UGV.

\subsection{Waypoint Graph}
\label{sec:appendix-waypoints}

\subsubsection{Common Principles}
\label{sec:appendix-waypoints-principle}

Both the UAV and the UGV operate on a predetermined discrete waypoint graph fixed at episode-set construction. Every movement action connects adjacent graph nodes. Movement cost and virtual time follow Section~\ref{sec:appendix-actions}.

\subsubsection{UAV Waypoint Grid}
\label{sec:appendix-waypoints-uav}

The UAV lays a two-dimensional grid with 40\,m spacing over the horizontal plane at 80\,m above the city. Only grid points within 40\,m of public roads are retained. Each episode activates points inside its search region and marks points outside as unreachable. A UAV move connects adjacent grid points in at most eight directions within the surrounding $3\times3$ neighborhood.

\subsubsection{UGV Road Network}
\label{sec:appendix-waypoints-ugv}

The UGV uses a sparse directed road network built from the public topology. Decision points occur about every 10\,m on straight segments and at intersection entries and exits, forks, and dead ends. This fixed map asset is shared by all episodes in the same town. Movement follows directed edges, and mileage accumulates along the traversed path as $\ell_e$ in SPL (Section~\ref{sec:appendix-metrics}).

\subsubsection{Runtime Options and Consistency}
\label{sec:appendix-waypoints-runtime}

At each decision step, legal motion candidates come from the fixed graph. The UAV receives at most eight reachable neighboring grid waypoints (Section~\ref{sec:appendix-waypoints-uav}). The UGV receives forward, turning, and reversing candidates from the road network (Section~\ref{sec:appendix-actions}).

\subsection{Evaluation Metrics}
\label{sec:appendix-metrics}

\subsubsection{Notation and General Scoring Conventions}
\label{sec:appendix-metrics-notation}

Let the evaluation set be $\mathcal{E}=\{1,\dots,N\}$. We assign each episode to success, method-side failure, or facility fault. Method-side failures comprise wrong-target confirmation, exhaustion of the task-step or logical-call budget, and method-side stalls. Facility faults are anomalies unrelated to the model or method. The decidable set $\mathcal{E}_{\mathrm{adj}}\subseteq\mathcal{E}$ contains successes and method-side failures, with $N_{\mathrm{adj}}=|\mathcal{E}_{\mathrm{adj}}|$. Unless stated otherwise, metrics are aggregated over $\mathcal{E}_{\mathrm{adj}}$, excluding facility faults. Let $S_e\in\{0,1\}$ be the success indicator. Aerial metrics use $\mathcal{E}_{\mathrm{aerial}}\subseteq\mathcal{E}_{\mathrm{adj}}$, the decidable episodes with an aerial search channel, with $N_{\mathrm{aerial}}=|\mathcal{E}_{\mathrm{aerial}}|$. The joint predicate ``distance, line of sight, and six-view visibility all hold'' is the non-verdict component of success in Eq.~\eqref{eq:confirm30} in the main paper. It requires the target to be visible in the UGV's six views, the horizontal distance not to exceed 30\,m, and line of sight to hold. The MATCH verdict required for task success is evaluated separately.

\subsubsection{Success Rate (SR)}
\label{sec:appendix-metrics-sr}

Success Rate measures whether the closed-loop task is completed:
\begin{equation}
\mathrm{SR}=\frac{1}{N_{\mathrm{adj}}}\sum_{e\in\mathcal{E}_{\mathrm{adj}}}S_e=\frac{\#\,\text{successes}}{N_{\mathrm{adj}}}.
\label{eq:sr}
\end{equation}

\subsubsection{Oracle Success Rate (OSR)}
\label{sec:appendix-metrics-osr}

Oracle Success Rate measures whether the team ever brings the UGV to a legal confirmation position. Let $O_e\in\{0,1\}$ indicate whether any pre-decision six-view snapshot in episode $e$ satisfies the joint predicate of Section~\ref{sec:appendix-metrics-notation}:
\begin{equation}
\mathrm{OSR}=\frac{1}{N_{\mathrm{adj}}}\sum_{e\in\mathcal{E}_{\mathrm{adj}}}O_e.
\label{eq:osr}
\end{equation}
OSR records a legal ground confirmation opportunity regardless of the final verdict. The OSR--SR gap counts episodes that reach such a pose and end without a correct confirmation, expressed as a fraction of all decidable episodes.

\subsubsection{Success weighted by Path Length (SPL)}
\label{sec:appendix-metrics-spl}

Let $\ell_e$ be the UGV's traveled distance in episode $e$, accumulated along road waypoints in meters. Let $\ell_e^{*}$ be the shortest legal confirmation route from an independent navigation oracle over the public road topology. For $\ell_e^{*}>0$, the per-episode contribution is
\begin{equation}
\mathrm{SPL}_e=S_e\cdot\frac{\ell_e^{*}}{\max(\ell_e,\ell_e^{*})},
\label{eq:spl-e}
\end{equation}
Failed episodes contribute zero. For a successful episode with $\ell_e^{*}=0$, the contribution is one when $\ell_e=0$ and zero otherwise. The aggregate is
\begin{equation}
\mathrm{SPL}=\frac{1}{N_{\mathrm{adj}}}\sum_{e\in\mathcal{E}_{\mathrm{adj}}}\mathrm{SPL}_e.
\label{eq:spl}
\end{equation}
SPL is reported only when all decidable episodes use one oracle version and have valid path measurements. In particular, a successful route shorter than the oracle route invalidates the aggregate rather than being discarded. We report SPL in percent.

\subsubsection{Navigation Error (NE)}
\label{sec:appendix-metrics-ne}

Navigation Error measures the remaining distance between the UGV and the target when navigation ends. Let $p_e^{\mathrm{UGV}}$ and $p_e^{o}$ be the UGV position and the target position (horizontal-plane coordinates) at the end of episode $e$:
\begin{equation}
\mathrm{NE}_e=\bigl\|p_e^{\mathrm{UGV}}-p_e^{o}\bigr\|_2,\qquad
\mathrm{NE}=\frac{1}{N_{\mathrm{adj}}}\sum_{e\in\mathcal{E}_{\mathrm{adj}}}\mathrm{NE}_e.
\label{eq:ne}
\end{equation}
NE includes both successful and failed episodes. It is reported only when finite terminal coordinates are available for every decidable episode.

\subsubsection{Aerial Search Efficiency (ASE)}
\label{sec:appendix-metrics-ase}

Let $T=80$ be the fixed task budget used for normalization. Let $\tau_e$ be the 1-based index in the UAV-local decision sequence at which the pre-decision frame first passes the oracle visibility probe, requiring target visibility and line of sight:
\begin{equation}
\mathrm{ASE}_e=
\begin{cases}
\dfrac{T-\tau_e+1}{T}, & \text{if a legal aerial observation occurs},\\[6pt]
0, & \text{otherwise},
\end{cases}
\label{eq:ase-e}
\end{equation}
\begin{equation}
\mathrm{ASE}=\frac{1}{N_{\mathrm{aerial}}}\sum_{e\in\mathcal{E}_{\mathrm{aerial}}}\mathrm{ASE}_e.
\label{eq:ase}
\end{equation}
ASE is computed only when aerial sampling is valid for every episode in $\mathcal{E}_{\mathrm{aerial}}$. It measures how early the UAV obtains a legal view of the target, not the spatial accuracy of the subsequent candidate. We report ASE in percent.

\subsubsection{Aerial Candidate Guidance (ACG)}
\label{sec:appendix-metrics-acg}

For the $j$-th candidate in episode $e$, ordered by creation time, let $d_{e,j}$ be the Euclidean distance from the target center to the nearest point of the candidate's evaluation square. The square has a half-side length of 15\,m and is used only for evaluation; the distance is zero when the target lies inside. The spatial quality and rank discount are
\begin{equation}
q_{e,j}=\frac{s}{s+d_{e,j}},\qquad w_j=\frac{1}{\log_2(j+1)},\qquad s=50\,\text{m},
\label{eq:acg-q}
\end{equation}
where $s$ is a distance scale. The spatial quality is one for a target inside the square and falls to one half at a distance of 50\,m from the square. Ground confirmation uses the horizontal UGV-to-target distance, with an upper limit of 30\,m. The episode-level score is the maximum discounted quality over all candidates, or zero when there are no candidates:
\begin{equation}
\mathrm{ACG}_e=\max_{j}\,w_j q_{e,j},\qquad
\mathrm{ACG}=\frac{1}{|\mathcal{E}_{\mathrm{aerial}}|}\sum_{e\in\mathcal{E}_{\mathrm{aerial}}}\mathrm{ACG}_e.
\label{eq:acg}
\end{equation}
Methods without an aerial candidate channel are marked ``--''. The rank discount favors accurate candidates that appear earlier. We report ACG in percent.

\subsubsection{Ground Verification Accuracy (GVA)}
\label{sec:appendix-metrics-gva}

A GVA event is a framework-executed CONFIRM or REJECT action at the ground confirmation site following an MLLM verdict. The target-centered oracle uses the non-verdict component of task success: a 30\,m confirmation circle, line of sight, and six-view visibility. NON\_MATCH and UNKNOWN both lead to REJECT. Table~\ref{tab:gva-support} cross-tabulates the executed action and the oracle label. Let $N_{\mathrm{event}}$ be the number of valid events:
\begin{equation}
\mathrm{GVA}=\frac{\mathrm{TP}+\mathrm{TN}}{N_{\mathrm{event}}}.
\label{eq:gva}
\end{equation}
GVA requires a valid oracle label for each event and valid sampling in every episode. Table~\ref{tab:gva-support} gives the number of contributing episodes and events for each configuration. Event counts range from 108 to 567 across the nine paired backbones. The oracle scores the verdict against target presence at the decision pose. The model returns a verdict for the six-view observation as a whole.

\begin{table*}[!t]
  \centering
  \footnotesize
  \setlength{\tabcolsep}{3pt}
  \caption{Ground verification events for the main backbones, fine-tuning comparisons, and the framework-comparison control. For an oracle-positive event (target within 30\,m of the UGV, clear line of sight, and target visible in the six-view observation), CONFIRM is a true positive (TP) and REJECT a false negative (FN). For an oracle-negative event, CONFIRM is a false positive (FP) and REJECT a true negative (TN). The Episodes column counts episodes contributing at least one event. GVA is $100(\mathrm{TP}+\mathrm{TN})/\mathrm{Events}$. Multiple events can come from the same episode. The Unseen-Test300 zero-shot row also appears under Qwen3.5-4B / AGOS-Agent.}
  \label{tab:gva-support}
  \begin{tabular}{@{}lccccccc@{}}
    \toprule
    Configuration & Events & Episodes & TP & TN & FP & FN & GVA (\%) \\
    \midrule
    \multicolumn{8}{@{}l}{\emph{Main paired backbones}} \\
    Qwen3.5-4B / AGOS-Agent & 567 & 284 & 113 & 412 & 25 & 17 & 92.59 \\
    Qwen3.5-4B / Baseline-Agent & 565 & 287 & 92 & 426 & 28 & 19 & 91.68 \\
    Qwen3-VL-4B / AGOS-Agent & 183 & 161 & 84 & 51 & 20 & 28 & 73.77 \\
    Qwen3-VL-4B / Baseline-Agent & 156 & 142 & 79 & 47 & 13 & 17 & 80.77 \\
    Qwen3.5-2B / AGOS-Agent & 439 & 250 & 56 & 330 & 29 & 24 & 87.93 \\
    Qwen3.5-2B / Baseline-Agent & 411 & 233 & 47 & 314 & 27 & 23 & 87.83 \\
    InternVL3\_5-4B / AGOS-Agent & 239 & 173 & 44 & 174 & 10 & 11 & 91.21 \\
    InternVL3\_5-4B / Baseline-Agent & 252 & 200 & 37 & 182 & 13 & 20 & 86.90 \\
    Gemma-3-4B / AGOS-Agent & 108 & 107 & 19 & 1 & 88 & 0 & 18.52 \\
    Gemma-3-4B / Baseline-Agent & 127 & 127 & 29 & 0 & 98 & 0 & 22.83 \\
    InternVL3-2B / AGOS-Agent & 120 & 120 & 19 & 0 & 101 & 0 & 15.83 \\
    InternVL3-2B / Baseline-Agent & 114 & 114 & 18 & 0 & 96 & 0 & 15.79 \\
    GPT-6 Astra / AGOS-Agent & 221 & 205 & 157 & 39 & 13 & 12 & 88.69 \\
    GPT-6 Astra / Baseline-Agent & 218 & 201 & 147 & 49 & 11 & 11 & 89.91 \\
    GLM-5.3 Flash / AGOS-Agent & 251 & 214 & 133 & 84 & 15 & 19 & 86.45 \\
    GLM-5.3 Flash / Baseline-Agent & 251 & 212 & 122 & 99 & 17 & 13 & 88.05 \\
    DeepSeek-4.1 Flash / AGOS-Agent & 246 & 202 & 112 & 105 & 21 & 8 & 88.21 \\
    DeepSeek-4.1 Flash / Baseline-Agent & 261 & 213 & 114 & 113 & 18 & 16 & 86.97 \\
    \midrule
    \multicolumn{8}{@{}l}{\emph{Fine-tuning comparisons}} \\
    Unseen-Test300 / Zero-shot & 567 & 284 & 113 & 412 & 25 & 17 & 92.59 \\
    Unseen-Test300 / Joint training & 222 & 221 & 171 & 3 & 48 & 0 & 78.38 \\
    Unseen-Test300 / Role-split training & 223 & 223 & 170 & 0 & 53 & 0 & 76.23 \\
    Seen-Test450 / Zero-shot & 786 & 383 & 127 & 576 & 44 & 39 & 89.44 \\
    Seen-Test450 / Joint training & 313 & 313 & 234 & 0 & 79 & 0 & 74.76 \\
    Seen-Test450 / Role-split training & 325 & 325 & 225 & 0 & 100 & 0 & 69.23 \\
    \midrule
    \multicolumn{8}{@{}l}{\emph{Framework-comparison control (Unseen-Test300)}} \\
    Qwen3.5-4B / Minimal-Agent & 71 & 71 & 46 & 0 & 25 & 0 & 64.79 \\
    \bottomrule
  \end{tabular}
\end{table*}

\subsubsection{Decision Steps (DS)}
\label{sec:appendix-metrics-ds}

Decision Steps is the mean number of MLLM decision calls per decidable episode. Configurations without MLLM participation are marked ``--''. Let $K_e=K_e^{\mathrm{UAV}}+K_e^{\mathrm{UGV}}$ count MLLM decision calls across the two roles. For AGOS-Agent and Baseline-Agent, each primary decision, crop verification, and final ground verification call counts once, including a call that leads to a legal fallback. More than one call can occur at a single task step. Algorithmic motion and transport-layer retries do not add decision steps. DS and its per-role decomposition $\mathrm{DS}_{r}$ are
\begin{equation}
\begin{aligned}
\mathrm{DS}&=\frac{1}{N_{\mathrm{adj}}}\sum_{e\in\mathcal{E}_{\mathrm{adj}}}K_e,\\
\mathrm{DS}_{r}&=\frac{1}{N_{\mathrm{adj}}}\sum_{e\in\mathcal{E}_{\mathrm{adj}}}K_e^{\,r},\quad r\in\{\mathrm{UAV},\mathrm{UGV}\},
\end{aligned}
\label{eq:ds}
\end{equation}

\subsubsection{Decision-Step-Weighted Success Rate (DSR)}
\label{sec:appendix-metrics-dss}

DSR is the counterpart of SPL along the decision dimension, further discounting success by the reciprocal of the number of decision steps:
\begin{equation}
\mathrm{DSR}=\frac{1}{N_{\mathrm{adj}}}\sum_{e\in\mathcal{E}_{\mathrm{adj}}}\frac{S_e}{\max(K_e,1)}.
\label{eq:dss}
\end{equation}
For configurations with MLLM participation, every successful episode has $K_e\geq1$, so $\mathrm{DSR}=\mathrm{SR}\cdot\mathbb{E}[\,1/K\mid\text{success}\,]$. We report DSR in percent. For configurations without MLLM participation, both DS and DSR are inapplicable and are marked ``--''.

\section{Dataset Construction}
\label{sec:appendix-dataset}

\subsection{Maps, Splits, and Distractors}
\label{sec:appendix-maps}

\begin{table}[t]
  \centering
  \footnotesize
  \caption{Map extent and episode distribution. Extent gives the $x$ and $y$ spans of the frozen road nodes, rounded to the nearest meter. For seen towns, the last column lists train, validation, and seen-test counts. Town03 is held out for unseen testing.}
  \label{tab:map}
  \setlength{\tabcolsep}{2pt}
  \resizebox{\columnwidth}{!}{%
  \begin{tabular}{@{}lcccc@{}}
    \toprule
    Town & Extent (m) & Episodes & Easy/Mid/Hard & Split counts \\
    \midrule
    Town01 & $398\times333$ & 505 & 155/175/175 & 300+75+130 \\
    Town04 & $929\times832$ & 505 & 155/175/175 & 300+75+130 \\
    Town05 & $487\times417$ & 505 & 155/175/175 & 300+75+130 \\
    Town02 & $201\times201$ & 155 & 155/0/0 & 100+25+30 \\
    Town10HD & $225\times210$ & 155 & 155/0/0 & 100+25+30 \\
    \midrule
    Town03 & $398\times417$ & 300 & 100/100/100 & 300 unseen \\
    \bottomrule
  \end{tabular}}
\end{table}

The training set uses 11 map--difficulty cells with 100 episodes per cell (Table~\ref{tab:map}). We held out Town03 because its road network accommodates the search regions for all three tiers and we judged its layout moderately challenging. The five seen towns each contribute 30 Easy episodes to Seen-Test450. Town01, Town04, and Town05 each contribute 50 Mid and 50 Hard episodes. Both test splits therefore have equal proportions of Easy, Mid, and Hard episodes. The seen split includes larger towns with more complex layouts, such as Town04.

Seen-Test450 was more challenging overall for the three configurations evaluated on both splits. Zero-shot, jointly trained, and role-split AGOS-Agent each achieved lower overall SR on Seen-Test450 (Table~\ref{tab:sftfull}).

Each Hard task includes two to three look-alike distractors. Selection prioritizes the same model in different colors, then a different model in the same color, and finally other body types. Among individual distractor vehicles in Hard Unseen-Test300, these categories account for 73.6\%, 24.0\%, and 2.4\%, respectively.

The catalog contains 25 recolorable vehicle models and nine models with fixed factory colors or liveries. Rendering the former in six standard colors gives $25\times6+9=159$ appearances across 34 models (main-paper Fig.~\ref{fig:dataset}(b)).

\subsection{Expert Collection and Annotation}
\label{sec:appendix-corpus}

\textbf{Collection timing.} Train Examples follow the online decision points described in Appendix~\ref{sec:appendix-verdicts}. The expert plans motion using public geometry, while ground truth supervises candidate nomination, acceptance, and confirmation. Model inputs use the online observation and action interfaces. Frames and action records for both members are included in each released Expert Trajectory.

\textbf{Training interfaces.} Action-decision answers are flat action objects with waypoint, candidate, or control parameters. UAV crop-selection answers contain a selected candidate and a target verdict, labeled as MATCH or UNKNOWN (insufficient visual evidence) in expert supervision. UGV final-verdict answers use the dedicated confirmation prompt of Section~\ref{sec:appendix-prompts-confirm}, which returns MATCH, NON\_MATCH, or UNKNOWN. Table~\ref{tab:verdicts} gives the runtime action for each verdict at aerial nomination and ground confirmation. Each Train Example follows the corresponding online prompt format and contains two to three ordered images and one structured executable answer.

\textbf{Category assignment.} A UAV primary-decision Train Example joins Verification when its episode, role, and task step also contain a crop-selection Train Example; the remaining UAV primary decisions are Move Decision. UGV candidate acceptance, arrival observation, release of an unverified candidate, and explicit confirmation actions join the final-verdict Train Examples in Confirmation. Other UGV search and recovery decisions are Move Decision. Categories are assigned before sequence-length filtering.

\textbf{Labeling rules.} At ground confirmation, a candidate whose center is more than 30\,m from the target receives NON\_MATCH. MATCH requires the target to be within 20\,m of the UGV and visible in at least one view. Other cases emit no training label. At crop verification, a box containing the projected target pixel receives MATCH when its candidate world position is within 30\,m of the target. All other crops receive UNKNOWN, which denotes insufficient crop evidence rather than a confirmed mismatch. At candidate acceptance, a candidate within 30\,m of the target is accepted; other cases emit no label. Acceptance initiates UGV approach during collection, while motion planning continues to use public geometry. \textbf{Labeling radius.} Positive ground-confirmation labels use a 20\,m radius during collection. Evaluation uses a 30\,m radius.

\subsection{Quality Filtering and Train Example Counts}
\label{sec:appendix-corpus-quality}

\textbf{Quality gate.} Fixed rules assign labels automatically from ground truth. Each Train Example passes checks for label format, action legality, image integrity, annotation timing, evidence eligibility, execution consistency, and source provenance; failures are excluded. Expert Trajectories from the same scenario are assigned to the same split.

\textbf{Train Example counts.} The release contains 27{,}720 Train Examples from 1{,}100 Expert Trajectories: 15{,}214 Move Decision, 9{,}048 Verification, and 3{,}458 Confirmation. The fine-tuning input contains 8{,}212 examples from 301 trajectories, comprising 4{,}850, 2{,}362, and 1{,}000 in the same category order, with 7{,}039 UAV and 1{,}173 UGV examples. The evaluated adapters were trained on the filtered 6{,}014 examples and were not retrained after the final release audit. A 5{,}120-token budget excludes 2{,}198 overlong examples (26.8\%), leaving 6{,}014 for training: 3{,}822 Move Decision, 1{,}648 Verification, and 544 Confirmation. The retained role counts are 5{,}297 UAV and 717 UGV, including 188 dedicated ground final-verdict examples.

\subsection{Split Isolation and Release Artifacts}
\label{sec:appendix-isolation}

The training and validation task-plan hashes are disjoint. The five seen maps are shared across splits, and Town03 is reserved for Unseen-Test300. All splits use the same vehicle catalog.

The four task manifests are Train1100, Val275, Seen450, and Unseen300. We plan to release the manifests, Expert Trajectories, Train Examples, content-addressed images and their manifest, configurations, framework and infrastructure code, and statistical source data. Five anonymous CSV files accompany this supplement and report paired comparisons, continuous-metric intervals, and GVA event counts.

We compared both Train1100 and the 301-episode fine-tuning input with Seen-Test450 and Unseen-Test300. These four comparisons share neither episode keys nor task-plan hashes. The two test sets also have disjoint episode keys and task-plan hashes. These checks establish disjointness by exact task identity. Semantic similarity and validation-set usage history fall outside their scope.

\section{AGOS-Agent Implementation}
\label{sec:appendix-agent}

\subsection{Algorithm-and-Tool Layer}
\label{sec:appendix-algorithm-tools}

\subsubsection{Joint Coverage}
\label{sec:appendix-coverage}

\textbf{Coverage geometry.} Coverage units are road waypoints within the region. A sampled UAV observation station covers units within its configured 25\,m sampling footprint. This effective planning radius is distinct from the camera's optical field-of-view footprint. The UGV records coverage at road waypoints it physically reaches and samples.

\textbf{Joint coverage.} Units coverable only from the air are first assigned to the UAV. The remaining units are allocated greedily to minimize the maximum estimated completion time across the two members, using incremental flight distance and directed road distance. The UAV visits observation stations in a sweep order, while the UGV follows legal shortest road routes. Reallocation preserves completed coverage and recomputes assignments from the current public positions. The coverage log records planned, arrived, and successfully sampled states separately from line-of-sight visibility.

\subsubsection{Candidate Projection}
\label{sec:appendix-projection}

After crop verification, the image coordinates of the candidate are back-projected to world coordinates and snapped to the nearest road waypoint:
\begin{equation}
\begin{split}
\mathbf{p}_{\text{world}} &= \begin{bmatrix} x_t^{\text{A}} \\ y_t^{\text{A}} \\ 0 \end{bmatrix} + h \cdot \mathbf{R}_{\text{cam}} \mathbf{K}^{-1} \begin{bmatrix} u_c \\ v_c \\ 1 \end{bmatrix},\\
\mathbf{v}_{\text{cand}} &= \arg\min_{\mathbf{v} \in \mathcal{V}_{\text{road}}} \big\| \mathbf{v} - \mathbf{p}_{\text{world}, xy} \big\|_2.
\end{split}
\end{equation}
Here $(u_c,v_c)$ are the candidate's pixel coordinates, and $(x_t^{\mathrm{A}},y_t^{\mathrm{A}})$ is the UAV's horizontal position at capture. The flight height is $h$, $\mathbf{R}_{\text{cam}}$ and $K$ are the camera rotation and intrinsics, and $\mathcal{V}_{\text{road}}$ is the road waypoint set. Back-projection assumes a fixed nadir-facing camera and a horizontal ground plane. The waypoint snap uses only $\mathbf{p}_{\text{world},xy}$.

\subsubsection{Event-Driven Wakeup}
\label{sec:appendix-wakeup}

\textbf{Wakeup priority.} Events have the following priority: unhandled requests, candidate arrival, supplementary-evidence requests during arrival or verification, new tool faults, and new detections above the similarity threshold. Algorithms route approach motion between event-triggered calls. Each role also has one routine check per episode when coverage progress passes the halfway point. Event-triggered calls do not count against this routine-check limit. The UAV reference-similarity evidence gate is
\begin{equation}
\begin{aligned}
s_t(b) &= \frac{1+\cos\!\big(\phi(O_t^{\mathrm{UAV}}[b]),\,\phi(I_{\text{ref}})\big)}{2},\\
g_t^{\mathrm{UAV}} &= \varmathbb{1}\!\Big[\exists b\in\mathcal{B}_t^{\mathrm{UAV}}:\\
&\qquad s_t(b)\ge\tau_{\text{wake}}\;\land\;\kappa_t(b)\notin\mathcal{M}_{\text{dedup}}\Big],
\end{aligned}
\end{equation}
where $\mathcal{B}_t^{\mathrm{UAV}}$ is the UAV proposal set, $\kappa_t(b)$ is the discretized evidence key, and $\tau_{\text{wake}}=0.55$.

\textbf{Perception and duplicate suppression.} Grounding DINO detects candidates and SigLIP 2 scores their similarity to the reference. The MLLM judges the candidate across views. The UAV nominates candidates for handoff, while the UGV uses near-range detection and confirmation.

\textbf{Shared perception settings.} AGOS-Agent and Baseline-Agent use \path{IDEA-Research/grounding-dino-base} (revision {\footnotesize\path{12bdfa3120f3e7ec7b434d90674b3396eccf88eb}}) and \path{siglip2-base-patch16-224}. The detector returns coordinates in the original RGB image. The encoder receives the full reference image and box-clipped RGB crops. SAM is disabled. Both roles use box and text thresholds of 0.20 and 0.18 and NMS IoU of 0.50.

The UAV retains up to three proposals after NMS. The UGV keeps up to eight detections per panel before similarity ranking and displays up to three crops. Each evidence card adds 20\% context, scales the long side to 256 pixels, and pads the crop to $256\times256$ below a 32-pixel label bar. Duplicate suppression uses role, viewpoint, 20\,m position bins, 30$^{\circ}$ yaw bins, 16-pixel box-coordinate bins, and 0.05 similarity bands.

\subsection{MLLM Layer}
\label{sec:appendix-mllm}

\subsubsection{Call Interface}
\label{sec:appendix-prompts-interface}

MLLM calls share one interface: a prompt, a JSON-structured public-state input, and ordered images. The model returns one JSON object without chain of thought. Images occupy up to three slots: the target reference, the current UAV frame or UGV six-view collage, and an optional set of at most three numbered candidate cards. A shared decision prompt covers Existence Judgment, Request Acceptance, movement, and recovery. Instance Judgment asks which numbered candidate matches the reference. Final Ground Verification asks whether the candidate vehicle being confirmed matches the reference.

\subsubsection{Candidate Handling and Verdicts}
\label{sec:appendix-verdicts}

A handoff request is processed after the receiver completes its current action. Candidate arrival, new evidence, and recovery events provide subsequent decision points. The prompt templates are listed in Appendix~\ref{sec:appendix-prompts}.

\textbf{Verdicts and candidates.} The system maintains one active candidate. MATCH denotes the reference vehicle, NON\_MATCH a different vehicle, and UNKNOWN insufficient evidence. At aerial nomination, NON\_MATCH drops the candidate and UNKNOWN defers judgment. The verifier's selected proposal is preferred. If it is NONE or invalid, the framework falls back to the proposal or candidate identifier from the primary decision; publication proceeds only when one of these identifiers matches a displayed proposal. At ground confirmation, MATCH submits the terminal claim. NON\_MATCH and UNKNOWN execute REJECT, release the candidate, and resume coverage outside its 30\,m suppression neighborhood. Table~\ref{tab:verdicts} summarizes these rules. Exhausting the logical model-call budget during verification leaves the candidate unresolved and terminates the episode as a method-side failure.

\begin{table}[t]
  \centering
  \footnotesize
  \caption{Verdict semantics by judgment site. A MATCH at confirmation ends the episode, as success or as a false confirmation (Eq.~\eqref{eq:confirm30} in the main paper); rejection is non-terminal and resumes coverage outside the suppressed neighborhood of the rejected candidate.}
  \label{tab:verdicts}
  \begin{tabular}{@{}lp{158pt}@{}}
    \toprule
    Verdict & Behavior across the two sites \\
    \midrule
    MATCH & nomination published; at confirmation the claim is accepted and the episode ends \\
    NON\_MATCH & nomination dropped; at confirmation the claim is rejected, the candidate released, and search resumes \\
    UNKNOWN & nomination published with judgment deferred; at confirmation the claim is rejected, the candidate released, and search resumes \\
    \bottomrule
  \end{tabular}
\end{table}

\begin{table*}[!t]
  \centering
  \fontsize{8}{9.5}\selectfont
  \setlength{\tabcolsep}{6pt}
  \caption{Evaluation settings for the nine paired backbones. Context and output limits are in tokens. For provider-managed models, the provider determines the context length.}
  \label{tab:backbone-settings}
  {%
  \begin{tabular}{@{}lccll@{}}
    \toprule
    Backbone & Context (tokens) & Output cap & Thinking & Temperature \\
    \midrule
    Qwen3.5-4B & 16{,}384 & 1{,}024 & Off & 0 \\
    Qwen3-VL-4B & 16{,}384 & 1{,}024 & Off & 0 \\
    Qwen3.5-2B & 16{,}384 & 1{,}024 & Off & 0 \\
    InternVL3\_5-4B & 16{,}384 / 32{,}768 & 1{,}024 & Off; thinking-off template & 0 \\
    Gemma-3-4B & 16{,}384 / 32{,}768 & 1{,}024 & Off & 0 \\
    InternVL3-2B & 32{,}768 & 1{,}024 & Off & 0 \\
    GPT-6 Astra & Provider-managed & 16{,}384 & High & 0 requested; API fallback \\
    GLM-5.3 Flash & Provider-managed & 16{,}384 & High & 0 requested; API fallback \\
    DeepSeek-4.1 Flash & Provider-managed & 16{,}384 & High & 0; no fallback \\
    \bottomrule
  \end{tabular}%
  }
\end{table*}

\section{Reproducibility Details}
\label{sec:appendix-reproduction}

\subsection{Evaluation Settings}
\label{sec:appendix-settings}

\textbf{Budget, decoding, and hardware.} Each episode has 80 task steps and a logical model-call budget of 1{,}000. On-device runs use greedy decoding (temperature $=0$), seed 20260725, and a context length fixed within each run and capped at 32{,}768 tokens. Evaluation uses RTX PRO 6000-class GPUs, CARLA-Air v0.1.7, and vLLM 0.28.0. API runs use high reasoning and a 16{,}384-token output cap, with temperature zero where supported; comparisons across API and on-device backends are descriptive. Each run uses fixed task plans, waypoint graphs, and evaluation configurations.

Settings are listed in Table~\ref{tab:backbone-settings}. All configurations check the legal menu, role, waypoint or candidate identifier, and remaining budget. Where indicated, the client requests the provider minimum if the API rejects temperature zero.

\subsection{Training Configuration}
\label{sec:appendix-train}

Table~\ref{tab:trainfull} gives the configuration shared by the joint and role-split adapters.

\noindent\begin{minipage}{\columnwidth}
\centering
  \footnotesize
  \captionof{table}{Complete LoRA supervised fine-tuning configuration.}
  \label{tab:trainfull}
  \begin{tabular}{@{}l p{0.56\columnwidth}@{}}
    \toprule
    Item & Value \\
    \midrule
    Backbone and precision & Qwen3.5-4B, BF16, vision tower and projector frozen \\
    Adapter & LoRA, rank 8, $\alpha=16$, no dropout, 16 q/v modules per adapter (layers 3, 7, 11, 15, 19, 23, 27, 31) \\
    Trainable parameters & approx. 0.92\,M per adapter \\
    Optimizer & AdamW, lr $1\times10^{-4}$, weight decay 0, grad clip 1.0 \\
    Batch and memory & micro-batch 1, grad accumulation 2, gradient checkpointing \\
    Sequence budget & 5{,}120 tokens; overlong Train Examples are isolated, not truncated \\
    Loss & assistant JSON replies only (including the end token), normalized by valid tokens \\
    Training amount & cap 4{,}096 steps / 1 epoch, whichever comes first \\
    Actual steps & joint 3{,}007 / UAV 2{,}649 / UGV 359 \\
    Seed & 20261027 (single random seed) \\
    Hardware & single-node single-card RTX PRO 6000-class GPU \\
    \bottomrule
  \end{tabular}
\end{minipage}\par

\textbf{Fine-tuning input.} The planned release index links the 301 episode keys to the 8,212 input records and identifies the 6,014 examples retained by the 5,120-token limit. Both training strategies use the same examples. Role-split training assigns 5,297 to the UAV adapter and 717 to the UGV adapter.

\subsection{Adaptation Parameters}
\label{sec:appendix-adapted-settings}

Decision calls in both adapted systems use Qwen3.5-4B with greedy decoding, seed 20260725, a 1{,}024-token output cap, and thinking mode disabled. Each episode allows 80 task steps, 1{,}000 logical model calls, and one terminal CONFIRM; REJECT does not consume this confirmation. Decision calls and PilotLLM evaluations count toward the logical-call budget. For AeroDuo-OS, the reported call count includes PilotLLM region evaluations. Grounding DINO is counted as a perception tool. Both methods use an online visual map at 0.5\,m per pixel.

\textbf{AGC-VLN-OS.} Planning uses a single-frame projection when both route endpoints lie within its observed area, and the accumulated map otherwise. The ground planner outputs ten pixel waypoints, which are mapped to road routes with a 15\,m waypoint tolerance. Routes span at most 14 edges, with up to two planning attempts per candidate, map version, and ground state. Failed model plans use a deterministic corridor route under the same edge limit. Arrival opens a three-step verification window, and search considers up to three service nodes per candidate.

\textbf{AeroDuo-OS.} The online map uses up to five keyframes. PilotLLM uses Qwen2-VL-2B-Instruct with the official adapter and probability head. Grounding DINO uses the public \texttt{grounding-dino-base} checkpoint, with box and text thresholds of 0.20 and 0.18, NMS IoU 0.5, and at most three proposals per view. Region search uses a 30\,m radius, routes of up to 14 edges, five progress milestones, and three service nodes. A matched proposal can initiate up to six visual-approach moves.

\subsection{Prompt Templates}
\label{sec:appendix-prompts}

\newtcolorbox{promptbox}[1]{colback=gray!6,colframe=gray!55,boxrule=0.4pt,arc=1.5pt,left=5pt,right=5pt,top=3pt,bottom=3pt,title={#1},fonttitle=\footnotesize\bfseries,coltitle=black,colbacktitle=gray!20}

\subsubsection{Decision Sites and Crop Verification}
\label{sec:appendix-prompts-decision}

The shared decision prompt receives the current phase (coverage, acceptance, approach, arrival, verification, recovery) and call reason as public input fields. Other fields include pose, the legal action menu, a public-state summary, message history, and candidate metadata. The framework supplies the action menu at each decision step. Below are representative verbatim excerpts of the original English system prompts; ellipses mark omissions.
\begin{promptbox}{Decision prompt (SYSTEM, excerpt of the original text; ellipses mark omissions)}
\footnotesize
``The framework supplies task\_phase and call\_reason; do not predict or change them. \dots\ Uncertainty is UNKNOWN, not NON\_MATCH. \dots\ Output one JSON object only, with decision and action; no reasoning/CoT/prose.''
\end{promptbox}

After the UAV proposes a candidate, one crop verification call presents the numbered candidate cards and requires a two-key answer. The verifier's selection is preferred; if it is NONE or invalid, the framework checks the identifier supplied by the primary decision before publishing. Aerial crops provide low-resolution screening evidence, while the final instance verdict is reserved for ground confirmation.
\begin{promptbox}{Crop verification prompt (VERIFY\_SYSTEM, excerpt of the original text)}
\footnotesize
``Answer with one JSON object with exactly two keys and no wrappers, no decision/action envelope, no reasoning: \texttt{\{"selected\_proposal": "<shown proposal ID or NONE>",\ "target\_verdict": "MATCH\textbar NON\_MATCH\textbar UNKNOWN"\}}. Select MATCH only when the chosen crop plausibly shows the reference vehicle; when nothing shown matches, selected\_proposal is NONE.''
\end{promptbox}

\subsubsection{Dedicated Prompt for the Ground Confirmation Site}
\label{sec:appendix-prompts-confirm}

Final ground confirmation uses the target reference and the native $1440\times540$ six-view collage labeled by direction. The prompt specifies the stopped-candidate situation, images, evidence rules, and a single-key reply. It asks the model to compare vehicle color and type. A MATCH triggers confirmation. NON\_MATCH and UNKNOWN trigger REJECT and release the candidate.
\begin{promptbox}{Ground confirmation prompt (CONFIRM\_SYSTEM, excerpt of the original text; ellipses mark omissions)}
\footnotesize
``Image 1 is the target reference image. Image 2 is your current six-view panorama (FRONT LEFT / FRONT / FRONT RIGHT / BACK LEFT / BACK / BACK RIGHT) captured from where you stand now. Use only these images. A nearby look-alike is not a match: answer MATCH only when the visible vehicle matches the reference in colour AND vehicle type. If the reference vehicle is not visible or cannot be judged from these views, answer UNKNOWN --- do not use NON\_MATCH for a vehicle you simply cannot resolve. \dots\ Answer with one JSON object with exactly one key, no wrappers, no decision/action envelope, no reasoning: \texttt{\{"target\_verdict": "MATCH\textbar NON\_MATCH\textbar UNKNOWN"\}}.''
\end{promptbox}

Baseline-Agent uses the same prompts and interface. It uses fixed coverage and calls the MLLM at coverage decision points.

\subsubsection{Adapted-System Prompt Excerpts}
\label{sec:appendix-adapted-prompts}

Selected prompt passages are reproduced below. Placeholders are filled with the current public observations and map geometry.

\begin{promptbox}{AGC-VLN-OS aerial localization (verbatim template)}
\footnotesize
Locate the reference vehicle in the current downward RGB. If absent or uncertain return \texttt{\{"visible":false\}}; otherwise return \texttt{\{"visible":true,"point\_px":[x,y],"confidence":0.5\}}. Coordinates in actual pixels, width=\{w\} height=\{h\}. Fixed altitude, no vertical action. Target: \{target\}. Images: target reference, current RGB, teammate reference (if provided). Exclude your UGV teammate; the teammate template has no location/background information.
\end{promptbox}

\begin{promptbox}{AGC-VLN-OS ground planning (verbatim template)}
\footnotesize
Plan exactly 10 ordered pixel waypoints following visible/public road lines. First=CAR, last=SERVICE. HYP is an unverified visual estimate, not GT. Yellow straight line is NOT a drivable route. Gray RGB is unobserved. Image width=\{w\} height=\{h\}; markers=\{markers\}. Return \texttt{\{"path":[[x,y],...]\}}. The task rectangle bounds the search, but legal roads may detour outside it before returning inside. Do not cut across buildings; use public roads. Pixel waypoints must lie in the shown raster; the executable road route between them may temporarily leave it.
\end{promptbox}

\begin{promptbox}{AeroDuo-OS ground verification (verbatim note)}
\footnotesize
GROUND VERIFICATION: the six panels are your only distance cue. The reference vehicle can be much closer than it looks (a few metres may read as tens of metres). If the reference target is visible in ANY panel and plausibly within about 30 m (it occupies a clearly large image area), answer CONFIRM now instead of driving. Answer REJECT only when you are confident the target is absent around you. Do not repeat STOP\_OBSERVE more than twice in a row; when the panels are unclear, keep searching with a movement action or the proposed visual approach.
\end{promptbox}

\begin{promptbox}{PilotLLM probability prediction (verbatim template)}
\footnotesize
Task: Predict the probability distribution of the drone's future flight locations to search for the target. Input Image: The image is an orthophoto map generated from the drone's past flight trajectory.
The top of the image corresponds to the north in the world coordinate system.
Target Information: The description of the target and its surrounding is shown below: \{target\}
Objective: Analyze the provided orthophoto map and target information. Predict the next flight location for the drone that maximize the probability of finding the target. Output a probability map, indicating the likelihood of different regions in the orthophoto map being the optimal next flight destinations.
\end{promptbox}

\end{document}